\documentclass[11pt]{article}
\usepackage[utf8]{inputenc}
\usepackage[letterpaper]{geometry} % enables changing of page margins
\usepackage{microtype}
\usepackage{graphicx}
\usepackage{subfigure}
\usepackage{booktabs} % for professional tables
\usepackage{Definitions}
\usepackage{enumitem}
\usepackage{algorithm}
\usepackage{algorithmic}
\usepackage{titling}
\usepackage[colorinlistoftodos]{todonotes}
\usepackage[colorlinks=true, allcolors=blue]{hyperref}
\usepackage[justification=centering]{caption}

\title{Representation Learning over Dynamic Graphs}
\author{Rakshit Trivedi\textsuperscript{\normalfont 1}, Mehrdad Farajtabar\textsuperscript{\normalfont 1}, Prasenjeet Biswal\textsuperscript{\normalfont 1}, \& Hongyuan Zha\textsuperscript{\normalfont 1} \\
\texttt{\{rstrivedi,mehrdad,pbiswal93\}@gatech.edu, zha@cc.gatech.edu}\\
\textsuperscript{1}College of Computing, Georgia Institute of Technology\\
}
\thanksmarkseries{arabic}
\date{}
\begin{document}
\maketitle
\begin{abstract}
How can we effectively encode evolving information over dynamic graphs into low-dimensional representations? In this paper, we propose \textbf{DyRep} -- an \textit{inductive} deep representation learning framework that learns a set of functions to efficiently produce low-dimensional node embeddings that evolves over time. The learned embeddings drive the dynamics of two key processes namely, \textit{communication} and  \textit{association} between nodes in dynamic graphs. These processes exhibit complex nonlinear dynamics that evolve at different time scales and subsequently contribute to the update of node embeddings. We employ a \textit{time-scale dependent} multivariate point process model to capture these dynamics. We devise an efficient unsupervised learning procedure and demonstrate that our approach significantly outperforms representative baselines on two real-world datasets for the problem of dynamic link prediction and event time prediction.

%
%realised through multi time-scale edges between nodes. We then propose an end-to-end 
%
%encode various information over temporal graphs where each edge has critical temporal information..
%
%two key dynamic processes over such temporal graphs association and communication -- 
%
%Representation Learning has emerged as a keystone task for many downstream machine learning applications. Specifically, automatic approaches leveraging deep learning techniques to encode
%structure in static graphs has gained extensive attraction. But many domains now involve dynamic graphs that contains critical temporal information
%over edges and interactions between nodes. This has risen compelling need for a principled approach that can encode graph information in such highly dynamic
%settings. To this end, we propose {\bf DyRep}, a deep representation learning framework that models multi-time scale dynamic processes of \textit{communication} and \textit{connection} observed over temporally evolving graphs. The observations are modeled as a multivariate point process whose intensity function is modulated by \textit{timescale} dependent function of these learned representations. 
\end{abstract}

\section{Introduction}
\label{intro}

Representation learning over graph structured data has emerged as keystone machine learning task due to its ubiquitous applicability in variety of domains such as social networks, bioinformatics, natural language processing, and relational knowledge bases. The key idea behind this task is to encode structural information at node (or subgraph) level 
into low-dimensional embedding vectors that can be used as feature inputs to further downstream tasks such as link prediction, clustering and classification~\cite{AkoTonKou15, GetTas07, LbKle03, SenNamBilMusGetGakEli08}. Traditionally, such domains have been modeled as static graphs where the learning is conducted on fixed set of nodes and edges~\cite{CaoLuXu15,GroLes16,PerAlSki14,TanQuWanZhaetal15,WanCuiZhu16,WanCuiWanPeiZhuYan17,XuWeiCaoYu17}. However, many of these domains now present data that is highly dynamic in nature. For instance, social network communications, financial transaction graphs, longitudinal citation data, etc., contain fine-grained temporal information on various components of evolving graphs. While research on learning representations over static graphs has progressed rapidly in recent years, there is a conspicuous lack of principled approach to tackle unique challenges involved in highly dynamic graph structures ~\cite{HamYinLes2017}.

Dynamic graphs have been typically treated from two viewpoints: a) Topologically evolving graphs where the number of nodes and edges are expected to grow (shrink) over time -- e.g. \textit{collaboration networks}. In this view, graphs are represented as collection of snapshots at discrete time points.  b) Temporal graphs where each edge contains temporal information -- e.g. \textit{telephone call networks}. Such graphs are represented as sequence of timestamped edges.
Most real-world dynamic graphs exhibit properties from both the above classes. We therefore take a unified abstract view of two classes and describe them as two temporal processes over a single dynamic graph --

{\bf Association Process:} Realized as growth of network and leads to long lasting information exchange between nodes. 

{\bf Communication Process:} Realized as interactions and leads to temporary information flow between nodes in graph.\\
At a fundamental level, the former constitutes the dynamics \underline{\textit{of}} the network which accounts for structural properties of network while the later constitutes the dynamics \underline{\textit{on}} the network which accounts for the activities and transmission properties of the network.~\cite{Far17, ArtRamMig17}.

We observe that such an abstraction naturally gives rise to a complex transmission system where information, contained in node's latent features, propagates across the graph in a nonlinear fashion. As noted in~~\cite{Cha12}, an important feature of such systems is the ability to express the dynamical processes at different scales. While it is natural to consider that the association and communication processes are directly observed and evolve at different temporal scales~\cite{FarWanGomLietal15}, we propose %that there exists 
an intermediate scale relevant to the hidden embedding propagation network that drives the interactions of the two processes across the graph. This leads to temporal evolution of nodes participating in these processes as the information ultimately propagates through them, updating their representations on the way. 

\textbf{Mediation Process:} We, therefore, posit \textit{representation learning over dynamic graphs} as the problem of learning node embeddings that serve as \textbf{mediator} influencing the dynamics of both the association and communication processes. On the other hand, both these complex nonlinear processes ultimately lead to evolution of nodes' representation. Further, the feature information propagated through the mediator nodes is governed by the temporal dynamics of communication and association histories of the node with it's neighbors in the graph.
%For instance, in a social network, frequent communication (likes or retweets) between users may drive association (friendship) and simultaneously association drives communication. 
\begin{figure*}[t]
\centering
\captionsetup{justification=justified}
\includegraphics[width = 0.9\textwidth]{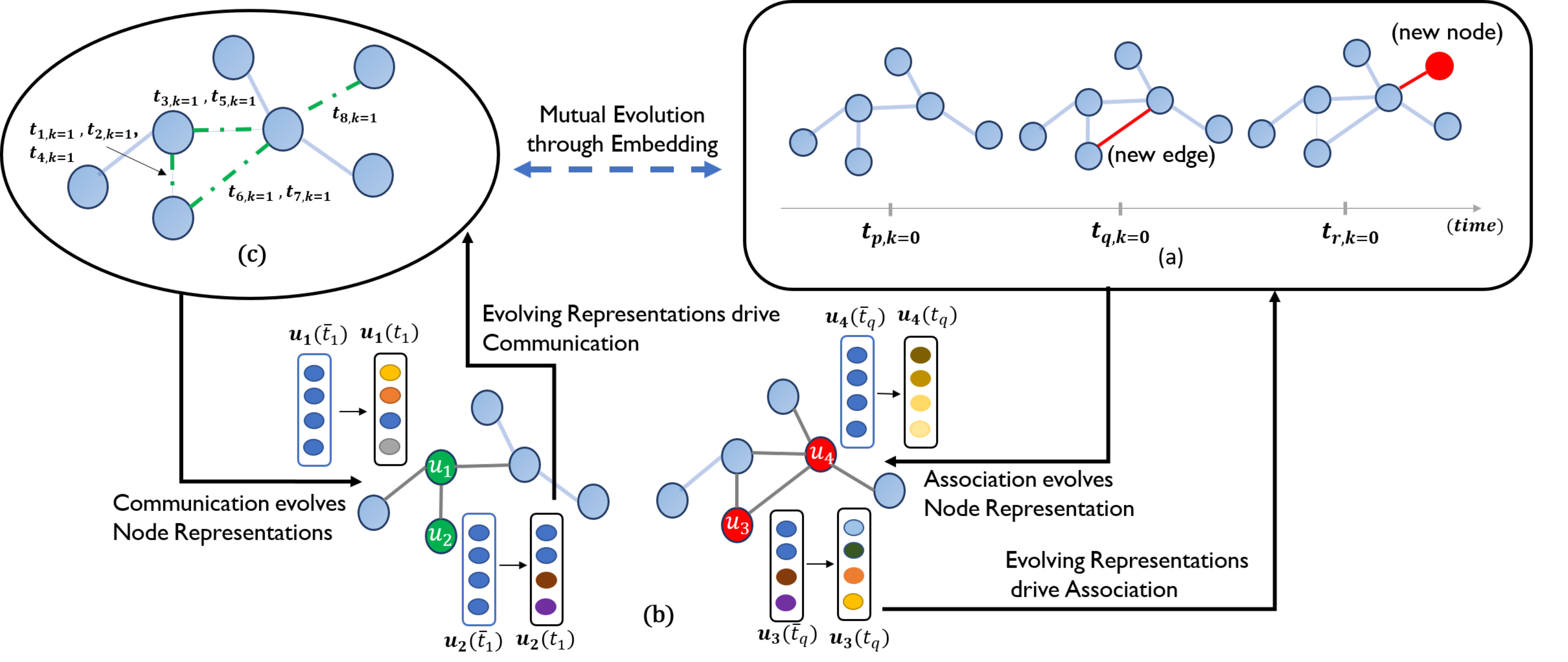}
\vspace{-4mm}
\caption{Illustrative View of DyRep Framework. \textbf{(a)} Association events where the node or edge grows (k=0, slow dynamics). \textbf{(c)} Communication Events where nodes interact with each other (k=1, fast dynamics). For both these processes, $t_{p,k=0} < (t_1, t_2, t_3, t_4, t_5)_{k=1} < t_{q,k=0} < (t_6, t_7)_{k=1} < t_{r,k=0}$. \textbf{(b)} Embedding Update Functions. Current illustration shows update of relevant node embeddings after first event of both kind. $\bar{t_1}$ and $\bar{t_q}$ are time points just before time $t_1$ and $t_q$.}
\label{fig:dyrep}
\vspace{-0.5cm}
\end{figure*}
For instance, in a social network, when a node's neighborhood grows, it changes node's social representation which in turn affects her social interactions (\textit{association} $\rightarrow$ \textbf{embedding} $\rightarrow$ \textit{communication}) and this effect is further percolated to all her neighbors (feature propagation). Similarly, when node's interaction behavior changes, it affects the representation of her neighbors and herself which in turn changes the structure and strength of her connections (\textit{communication} $\rightarrow$ \textbf{embedding} $\rightarrow$ \textit{association}). Thus, a node's evolving and latent representation play a mediator's role in driving the dynamics of the two observed processes. We call this phenomenon \textbf{ \textit{evolution through mediation}}. 

To model this phenomenon, we propose \textbf{DyRep}, that considers joint effect of communication and association processes for updating node representations and simultaneously models the driving effect of evolving representations on both these processes.   To achieve this, we identify key challenges involved in the learning process and make several contributions addressing these challenges:

Dynamic processes over graphs bring inherent \textbf{uncertainty} due to possibility of encountering multiple new nodes and edges over time. To address this, we propose an \emph{inductive} framework that learns set of functions to compute node embeddings instead of being restricted to learning only individual representations. We then equip our framework with mathematical benefits of temporal point processes to effectively model \textbf{fine-grained temporal dynamics}. Further, as the processes evolve at \textbf{multiple time scales}, we extend the point process model to consider a time scale dependent parameter.  To capture highly \textbf{complex and nonlinear dynamics} governing the evolution of node representations, we use a deep recurrent architecture with powerful structural encoder. A novel intensity controlled attention mechanism is proposed to facilitate the \textbf{mediation} process. Finally, dynamic graphs has potential to generate enormous amount of communication events and hence we propose an efficient learning procedure to achieve high \textbf{scalability} in number of events.
Figure~\ref{fig:dyrep} provides an illustrative view of our framework. We show the effectiveness of our model through both quantitative and exploratory analysis against several representative baselines on two real-world dynamic graphs.

\vspace{-2mm}
\section{Preliminaries}
We briefly describe the basic mathematical framework of temporal point process that is used to model temporal dynamics of both association and communication events. We then specify the dynamic graph setting that exhibits both these types of events.
\label{bkgrnd}
\vspace{-2mm}
\subsection{Temporal Point Process}
\vspace{-2mm}
Stochastic point processes~\cite{DalVer07} are random processes whose realization comprised of discrete events in time, $\cbr{t_1, t_2, \ldots}$. A temporal point process is one such stochastic process that can be equivalently re\-pre\-sen\-ted as a counting process, $N(t)$, which contains the number of events up to time $t$. 

The common way to characterize temporal point processes is via the conditional intensity function $\lambda(t)$, a stochastic model of rate of happening events given the previous events. Formally, $\lambda(t)\rd t$ is the conditional probability of observing an event in the tiny window $[t, t+\rd t)$,
$\lambda(t)\rd t := \PP\cbr{\text{event in }[t, t+\rd t)|\Tcal(t)} = \EE[\rd N(t) | \Tcal(t)]$, where 
$\Tcal(t)=\cbr{t_k|t_k<t}$ is history until $t$.
% \vspace{-0.2cm}
% \begin{equation}
% \begin{split}
% \lambda(t)\rd t &:= \PP\cbr{\text{event in }[t, t+\rd t)|\Tcal(t)} \\
% &= \EE[\rd N(t) | \Tcal(t)]
% \end{split} 
% \end{equation}
%where one typically assumes that only one event can happen in a small window of size $\rd t$, \ie,~$\rd N(t) \in \cbr{0,1}$. 

Similarly, for $t > t_n$ and given history $\Tcal=\cbr{t_1,\dotso, t_n}$,  we characterize the conditional probability that no event happens during $[t_n, t)$ as $ S(t|\Tcal) = \exp\left(-\int_{t_n}^{t} \lambda(\tau) \, \rd\tau \right)$, which is called  survival function of the process~~\cite{AalBorGje08}.
Moreover, the conditional density that an event occurs at time $t$ is defined as $
f(t) = \lambda(t)\, S(t)
$.
The intensity $\lambda(t)$ is often designed to capture phenomena of interests -- common forms include 
Poisson Process, Hawkes processes~\cite{FarDuRodValZhaSon14,Hawkes71,WanXieDuSon16,TabValFarSonSchGom17}, Self-Correcting Process~\cite{IshWes79}. Temporal Point Processes have previously been used to model both -- dynamics on the network~\cite{FarYeHarSonZha16,ZarKhoFarRabZha17,FarYanYeXuTriKhaLiSonZha17} and dynamics of the network~\cite{TraFarSonZha15,FarWanGomLietal15}.

\vspace{-2mm}
\subsection{Dynamic Graph Representation}
\vspace{-2mm}
Let $\mathcal{G}_{t_0} = (\mathcal{V}, \mathcal{E}_{t_0})$ be the initial snapshot of graph $\mathcal{G}$ at time $t_0 = 0$, where $\mathcal{V}$ is the set of nodes and $\mathcal{E}$ is the set of edges. Let $|\mathcal{V}| = n$ and $|\mathcal{E}| = m$. For this work, we assume that the topology evolves through growth and there is no deletion of edge or node which is intended as future work.

{\bf Event Observation.} Both communication and association processes are realized in the form of $\mathcal{D}$ dyadic events observed on graph $\mathcal{G}$ over a temporal window $[t_0,T)$. We use the following canonical tuple representation for any type of event at time $t$ of the form $e = (u, v, t, l, k)$,
where $u, v$ are the two nodes involved in an event. $t$ represents time of the event. $l \in \{0,1\}$ represent link status -- $l=1$ signify a association edge and $l=0$ signify no association. $k \in \{0,1\}$ represent the type of event. Here, $k=0$ when the current event is a association event and $k=1$ when the current event is a communication event. One can use $k>1$ to signify more types of communication events (e.g. in dynamic heterogeneous graphs). We then represent complete set of $P$ observed events ordered by time in window $[0,T]$ as $\mathcal{O} = \{(u, v, t, l, k)_p\}_{p=1}^P$ . Here, $t_p \in \mathbb{R}^+$, $0 \le t_p \le T$.\\\\
{\bf Node Representation.} Let $\mathbf{z}^v \in \mathbb{R}^d$ represent $d$-dimensional representation of node $v$. As the representations are evolving over time, we qualify the representation as function of time and  denote it as $\mathbf{z}^v(t)$ which signifies the representation of node $v$ updated after an event involving $v$ at time $t$. We use $\mathbf{z}^v(\bar{t})$ to denote most recently updated  embedding of node $v$ just before time $t$. 

\section{Proposed Method: DyRep}
\vspace{-2mm}
\label{model}

We propose an \textit{inductive} representation learning framework that learns a set of functions capable of ingesting
temporally evolving information about the nonlinear dynamics governing the changes in topological structure of graph and interactions between the nodes in the graph. These functions generate and/or update node embeddings over time based on the ingested information. The key idea behind our approach is that  when an event is observed between two nodes, information flows from the neighborhood of one node to the other and affects the the embeddings of the nodes accordingly. While a communication event only propagates local information across two nodes, an association event changes the topology and thereby has more global effect. To capture the mutual effects of changing graph topology and interaction between nodes while learning temporally evolving node representations, we design the following three functions:

 {\bf Temporal Function:} A multi time-scale conditional intensity function that models the occurrence of events and is learned through a single layer neural network over most recent embeddings of the nodes involved.

 {\bf Embedding Update Function:} A deep recurrent function to update the embedding of nodes involved in an event based on its own previous embedding, aggregate information passed from the neighborhood of the other node and any exogenous effect.
 
{\bf Attentive Aggregate Function:} An aggregator function to combine the information from the neighborhood of a node that serves as input to the aforementioned embedding update function. The information is aggregated by employing a novel intensity augmented attention scheme to attend to nodes according to past association or communication.

\vspace{-2mm}
\subsection{Temporal Function: Modeling Multi time-scale Global Dynamics}
\vspace{-2mm}

The observations over dynamic graph contain temporal point patterns of two interleaved complex processes in the form of communication and association events respectively. As mentioned before, both these process are dependent on the most recent node representations. To incorporate these complex processes in our model, we first formalize the notion of compatibility or closeness between representations. 
Given an observed event $p = (u, v, t, l, k)$, we define a function $g(\bar{t})$ over the most recently updated representations of two nodes, $\mathbf{z}^u(\bar{t})$ and $\mathbf{z}^v(\bar{t})$ that computes the compatibility between the two representations as follows:
%\begin{equation}\label{g1}
%g(\mathbf{z}^s(t-), \mathbf{z}^o(t-)) = \mathbf{z}^s(t-)^T \cdot \mathbf{z}^o(t-)
%\end{equation}
%\begin{center}
%or
%\end{center}
\begin{equation}\label{g2}
g_k(\bar{t}) = \bm{\omega}_k^T \cdot [\mathbf{z}^u(\bar{t});\mathbf{z}^v(\bar{t})]
\end{equation}
Here, $k$ is the type of event (communication vs. association) and hence $\bm{\omega}_k \in \mathbb{R}^{2d}$ serves as the model parameter that learns time-scale specific compatibility score. 

Using this notion of compatibility, we employ a continuous-time deep model of temporal point process and use the function in~(\ref{g2}) to parametrize the conditional intensity function $\lambda_k^{u,v}(t)$ that models the occurrence of event $p$ of type $k$ at time $t$:
\begin{equation}\label{intensity}
\lambda_k^{u,v}(t) = f_k(g_k(\bar{t}))
\end{equation}
The choice of $f_k$ needs to account for two critical criteria: 1.) The definition of $g(\cdot)$ allows the value of $g(\cdot)$ to be negative. But we require the intensity to be positive. 2.) As mentioned before, the dynamics corresponding to communication and association processes evolve at different time scales and hence we use a modified version of softplus function parameterized by a dynamics parameter $\psi_k$ to capture this \textit{timescale} dependence:
\begin{equation}\label{scale}
f_k(x) = \psi_k \log(1 + \exp(x/\psi_k))
\end{equation}
where, $x = g(\bar{t})$ in our case and $\psi_k$ is scalar time-scale parameter learned as part of training. In one-dimensional event sequences, this formulation in~(\ref{scale}) corresponds to the nonlinear transfer function proposed in~\cite{MeiEis17}

\vspace{-2mm}
\subsection{Embedding Update Function: Modeling Local Information Propagation Dynamics}
\vspace{-1mm}

In this section, we learn an embedding update function that models the nonlinear evolution of a given node's representation for both nodes involved in an observed event. Specifically, we propose that after a node has participated in an event, the representation of the node needs to be updated to capture the effect of the observed event based on the principles of - \emph{self-propagation}, \emph{localized feature information propagation through the other node} and \emph{exogenous drive}.

{\bf Self-Propagation.} Self-propagation can be considered as foundational component of the dynamics governing an individual node's evolution. A node evolves in embedded space with respect to its previous position (e.g. set of features) and do not evolve randomly. 

{\bf Exogenous Drive.} Some exogenous force may smoothly update the node's current features during the time interval between two global events involving that node. 

{\bf Localized Embedding Propagation.} Two nodes involved in an event form a temporary (communication) or a permanent (association) pathway for the information to propagate from the neighborhood of one node to the other node. This corresponds to the influence of the nodes at second-order proximity passing through the other node participating in the event. 
To realize the above processes in our setting, we first describe a simple setup: Consider nodes $u$ and $v$ participating in any type of event at time $t$. Let $\mathcal{N}_u$ and $\mathcal{N}_v$ denote the neighborhood of nodes $u$ and $v$ respectively. 

We discuss two key points here: 1.) Node $u$ serves as mediator passing information from $\mathcal{N}_u$ to node $v$ and hence $v$ receives the information in an aggregated form through $u$. 2.) While each neighbor of $u$ wants to pass its information  to $v$, the information that node $u$ captures is governed by an aggregate function parametrized by $u$'s communication and association history with its neighbors. 

With this setup, for any event at time $t$, we update the embeddings for both nodes involved in the event using a recurrent architecture. Specifically, for $p$-th event of node $v$, we evolve its representation with the following update function:
\begin{equation}\label{update_eq}
\vspace{-1mm}
\mathbf{z}^v(t_p) = \sigma(\underbrace{\mathbf{W}^{struct} \mathbf{h}^u_{struct}(\bar{t_p})}_\text{Localized Embedding Propagation}+ \underbrace{\mathbf{W}^{rec}\mathbf{z}^v(t_{p-1})}_\text{Self-Propagation} + \underbrace{\mathbf{W}^{t}(t_p-t_{p-1})}_\text{Exogenous Drive})
\vspace{-1mm}
\end{equation}
where, $\mathbf{h}^u_{struct} \in \mathbb{R}^d$ is the output representation vectors obtained from aggregator function on node $u$'s neighborhood and $\mathbf{z}^v(t_{p-1})$ is the recurrent state obtained from the previous representation of node $v$. $t_p - t_{p-1}$ is a scalar difference between the current and previous time points when the node $v$ was involved in an event. $\mathbf{W}^{struct}, \mathbf{W}^{rec} \in \mathbb{R}^{d \times d}$ and $\mathbf{W}^{t} \in \mathbb{R}^d$ are model parameters that govern the aggregate effect of all the three processes respectively. Eq.~(\ref{update_eq}) is used to make updates for both nodes involved in the event with all three components specific to the node. 
%$\mathbf{h_{struct}}$ and time interval
% \subsubsection{Learning $\mathbf{h_{rec}}$}
% \RT{This subsection 2.3.1 is obsolete. Just kept for check - will be removed soon.}
% For computing a node representation $\mathbf{z}(t_p)$ at time $t_p$, this vector corresponds to the previous state of $\mathbf{z}$ i.e. most recent representation. This component provides the recurrent state and hence can be used to model various changes in node representations over time.
% For simplicity, we currently define it as:
% \begin{equation}\label{rec}
% \mathbf{h_{rec}}(\bar{t_p}) = \mathbf{z}(\bar{t_p})
% \end{equation}
% The above equation will only consider the conventional dependency and changing interests of user over time as it contributes to the update of representation from the previous one. But one can extend this by either using attention mechanism or memory augmented recurrent unit like GRU to capture decay and peak in interest at finer level. 
\vspace{-2mm}
\subsection{Attentive Aggregate function: Modeling Mesoscopic Influence Dynamics}
\vspace{-2mm}
In a dynamic graph, any event would not only affect the nodes involved in that event but also the local neighbors of those nodes. Aligned with our motivation, any event would impact the embeddings of the nodes involved in those events and the updated embeddings would further induce more events across graph. To capture this mutual effect, we want to design a function that can capture the structural information local to a node based on historical events on its neighbor nodes. To this effect, we propose a novel \textit{Intensity based Attentional Mechanism} that empowers the aggregate function to attend to the neighbors based on node's communication and association history.

{\bf Intensity based Attentional Mechanism.} Let $\mathbf{A}(t) \in \mathbb{R}^{n \times n}$ be the adjacency matrix for graph $\mathcal{G}_t$ at time $t$. Let $\mathcal{S}(t) \in \mathbb{R}^{n \times n}$ be a stochastic matrix capturing the association strength between pair of vertices at time $t$. One can consider the matrix $\mathcal{S}$ as a selection matrix that induces a \textbf{\textit{natural selection}} process for a node -- it would tend to communicate more with other nodes that it wants to associate with or has recently associated with. On the other hand it would want to attend less to non-interesting nodes.
While we discuss construction and update of $\mathcal{S}$ in the next section, following implication is required for the construction of $\mathbf{h}^u_{struct}$ in~(\ref{update_eq}): For any two nodes $u$ and $v$ at time $t$,  $\mathcal{S}_{uv}(t) \in [0,1]$ if $\mathbf{A}_{uv}(t) = 1$ and $\mathcal{S}_{uv}(t) = 0$ if $\mathbf{A}_{uv}(t) = 0$. Denote $\mathcal{N}_u(t) = \{i: \mathbf{A}_{iu}(t) = 1\}$ as the 1-hop neighborhood of node $u$ at time $t$.\\\\
To formally capture the difference in the influence of different neighbors, we design a novel \textit{intensity based attention layer} that uses the matrix $\mathcal{S}$ to induce a shared attention mechanism to compute normalized attention coefficients for nodes. Specifically, we inject the graph structure by performing \emph{localized attention} - for a node $u$, compute the coefficients pertaining to the 1-hop neighbors $i$ of node $u$ as follows:
\begin{equation}\label{attend}
q_{ui}(t) = \frac{\exp(\mathcal{S}_{ui}(\bar{t}))}{\sum_{i' \in \mathcal{N}_u(t)}\exp(\mathcal{S}_{ui'}(\bar{t}))}
\end{equation}

where $q_{vi}$ signify the attention weight for the neighbor $i$ at time $t$ and hence it is a temporally evolving quantity. These attention coefficients are 
then used to compute the aggregate information $\mathbf{h}_{struct}^u(\bar{t})$ for node $v$ by employing an attended aggregation mechanism across neighbors as follows: 
\vspace{-1mm}
\begin{equation*}\label{struct}
\mathbf{h}_{struct}^u(\bar{t}) = \max\left(\left\{\sigma\left(q_{ui}(t) \cdot\mathbf{h}^i(\bar{t})\right),\forall i \in  \mathcal{N}_u(\bar{t}) \right\}\right)~\text{where}~\mathbf{h}^i(\bar{t}) = \mathbf{W}^{h}\mathbf{z}^i(\bar{t}) + \mathbf{b}^{h}
\end{equation*}

where, $\mathbf{W}^{h} \in \mathbb{R}^{d \times d}$ and $\mathbf{b}^{h} \in \mathbb{R}^{d}$ are parameters governing the information propagated by each neighbor of $u$. $\mathbf{z}^i(\bar{t}) \in \mathbb{R}^d$ is the most recent embedding for node $i$. 

The attention mechanism plays two simultaneous roles: On one hand it captures the effect of contribution of each neighbors based on the intensity of events between them. On the other hand it can also be seen as inducing competition among neighbors to gain more attention of the node. 

Overall, this component captures information from the structure of the graph and associate it with the temporal dynamics of evolving representations. The intuition is that a node is highly influenced by its neighbors. While the effect may be propagated from far away nodes, it still needs to propagate through immediate neighbors. Most previous approaches either directly observe peer influence explicitly or use fixed length proximity measure to capture local information (many static graph embedding approaches). But evolving representations are expected to capture both temporal and structural dependencies through their second order proximity neighborhood and thereby account for hidden propagation across network. 

\begin{algorithm}[h]
   \caption{Update Algorithm for $\mathcal{S}$ and $\mathbf{A}$}
   \label{alg:alg1}
\begin{algorithmic}
   \STATE {\bfseries Input:} Event record $o = (u, v, t, l, k)$, Event Intensity $\lambda_k^{u,v}(t)$ computed in (\ref{intensity}), most recently updated $\mathbf{A}(t-)$ and $\mathcal{S}(t-)$.
   \STATE {\bfseries Output:} $\mathbf{A}(t)$ and $\mathcal{S}(t)$   
   \STATE   
   \STATE {\bf 1.} Update $\mathbf{A} : \mathbf{A}(t) = \mathbf{A}(t-)$
   \IF{$k = 0$} 
   \STATE $\mathbf{A}_{uv}(t) = \mathbf{A}_{vu}(t) = 1$ 
   \ENDIF
   \STATE
   \STATE {\bf 2.} Update $\mathcal{S} : \mathcal{S}(t) = \mathcal{S}(t-) $
   \IF {$k = 1$ and $l = 0$}
   \STATE \textbf{return} $\mathcal{S}(t), \mathbf{A}(t)$
   \ENDIF
   \FOR{$j \in \{u, v\}$}
   \STATE $b = \frac{1}{|\mathcal{N}_j(t)|}$ where $|\mathcal{N}_j(t)|$ is the size of $\mathcal{N}_j(t) = \{i: \mathbf{A}_{ij}(t) = 1\}$
   \STATE $\mathbf{z} \leftarrow \mathcal{S}_{j}(t)$
   %\STATE $\mathbf{v} \leftarrow \mathcal{S}_v$
   \IF {$k = 1$ and $l = 1$}
   \STATE $\mathbf{z}_i = b + \lambda_k^{ji}(t)$ where $i$ is the other node involved in the event.  
   \ELSIF {$k = 0$ and $l = 0$}
   \STATE $b' = \frac{1}{|\mathcal{N}_j(t-)|}$ where $|\mathcal{N}_j(t-)|$ is the size of $\mathcal{N}_j(t-) = \{i: \mathbf{A}_{ij}(t-) = 1\}$
   \STATE $x = b' - b$
   \STATE $\mathbf{z}_i = b + \lambda_k^{ji}(t)$ where $i$ is the other node involved in the event   
   \STATE $\mathbf{z}_w = \mathbf{z}_w -x$;~~ $\forall w \ne u$
   \ENDIF
   \STATE Normalize $\mathbf{z}$
   \STATE $\mathcal{S}_{j}(t) \leftarrow \mathbf{z}$
   \ENDFOR
   \STATE \textbf{return} $\mathcal{S}(t), \mathbf{A}(t)$
\end{algorithmic}
\end{algorithm}

{\bf Construction and Update of $\mathcal{S}$.} As mentioned earlier, while modeling the global and local principles governing the evolutionary processes, the key to effectively learn the overall dynamics of the system is an intermediate level of information propagation which connects the two vastly different scales of events. We construct a single stochastic matrix $\mathcal{S}$ (used to parameterize attention in the earlier section) to capture this intermediate information. We note that updates in adjacency matrix $\mathbf{A}$ over time and activities between pair of nodes directly effect the values in of node representations which in turn impact values in $\mathcal{S}$. And on the other way, updates made to $\mathcal{S}$ contribute further activities between nodes and eventual update to $\mathbf{A}$ through its effect on node representations.\\\\
As before, $\mathbf{A}(t) \in \mathbb{R}^{n \times n}$ and $\mathcal{S}(t) \in \mathbb{R}^{n \times n}$ are the adjacency matrix and a right-stochastic matrix at time $t$. At the initial timepoint $t = t_0$, we construct $\mathcal{S}(t_0)$ directly from $\mathbf{A}(t_0)$. Specifically, for a give node $v$, we initialize the elements of corresponding row vector $\mathbf{\mathcal{S}}_v(t_0)$ as:
\begin{equation}
\mathcal{S}_{vu}(t_0) = \begin{cases}
0 &\text{if $v=u$}\\
0 &\text{if $\mathbf{A}_{vu}(t_0) = 0$}\\
\frac{1}{|\mathcal{N}_v(t_0)|} &\text{$\mathcal{N}_v(t_0)= \{u: \mathbf{A}_{uv}(t_0) = 1\}$}
\end{cases}
\end{equation}
After observing an event $o = (u, v, t, l, k)$ at time $t > t_0$, we make updates to $\mathbf{A}$ and $\mathcal{S}$ as per the observation of $l$ and $k$. Algorithm~\ref{alg:alg1} describes the update scenarios.

\section{Efficient Learning Procedure}
\vspace{-2mm}

The complete parameter space for the current model is $\mathbf{\Omega} = \{\mathbf{V},\mathbf{W}^{struct},\mathbf{W}^{rec},\mathbf{W}^t,\mathbf{W}^{h}, \mathbf{b}^{h},\\ \{\bm{\omega}_k\}_{k=0,1},\{\psi_k\}_{k=0,1}\}$. For a set $\mathcal{O}$ of $P$ observed events, 
we learn these parameters by minimizing the negative log likelihood of the intensity function in~(\ref{intensity}):

\begin{equation}\label{loss}
\mathcal{L} = - \sum_{p=1}^P \log\left(\lambda_p(t)\right) + \int_{0}^T\Lambda(\tau)d\tau
\end{equation}
where $\lambda_p(t) = \lambda^{u_p,v_p}_{k_p}(t)$ represent the intensity of event at time $t$  and $$\Lambda(\tau) = \sum_{u=1}^n\sum_{v=1}^n\sum_{k \in \{0,1\}} \lambda^{u,v}_k(\tau)$$
represent total survival probability for events that do not happen.

\begin{figure*}[h]
\small
\centering
\begin{tabular}{c}
\includegraphics[width = 0.4\textwidth]{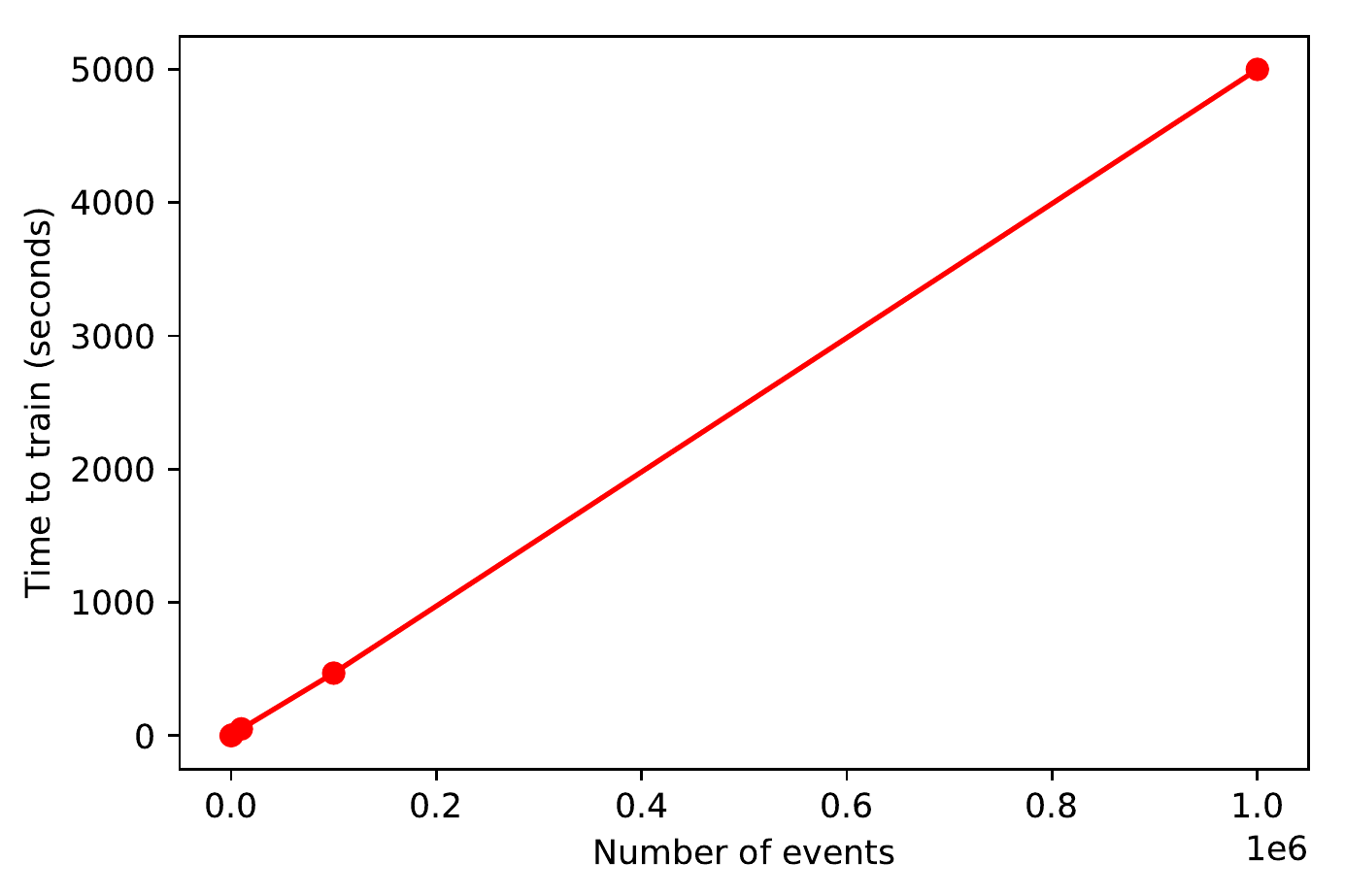}
\end{tabular}
\caption{Scalability in number events}
\vspace{-2mm}
\label{fig:scal}
\end{figure*}

While it is intractable (will require $\mathcal{O}(n^2k)$ time) and unnecessary to compute the integral in Eq. (\ref{loss}) for all possible non-events in a stochastic setting, we can locally optimize the objective function using mini-batch stochastic gradient descent where we estimate the integral using novel sampling technique. 
Algorithm~\ref{alg:alg2} provides the recipe to compute the survival term in Eq. (\ref{loss}). Let $m$ be the mini-batch size and $N$ be the number of samples. The complexity of Algorithm~\ref{alg:alg2} will then be $\mathcal{O}(2mkN)$ for the batch where the factor of $2$ accounts for the update happening for two nodes per event. Figure~\ref{fig:scal} shows the running time for training when the number events are in  $\{10^3, 10^4, 10^5, 10^6 \}$. It demonstrates linear scalability in number of events which is desired to tackle web-scale dynamic networks~\cite{ParBenLes17}.

\begin{algorithm}[t!]
   \caption{Computation of integral term in Eq.~\ref{loss} for a mini-batch}
   \label{alg:alg2}
\begin{algorithmic}
   \STATE {\bfseries Input:} Minibatch $\mathcal{M} = \{m_q = (u, v, t, l, k)_q\}_{q=1}^{|\mathcal{M}|}$. Minibatch node list $\mathbf{l}$, sample size $N$.
   \STATE {\bfseries Output:} Minibatch survival loss $L_{surv}$
   \STATE
   \STATE $L_{surv}=0.0$
   \FOR{$q = 0~to~|\mathcal{M}| - 1$}
   \STATE $t_{curr} = m_q \rightarrow t$
   \STATE $u_{curr} = m_q \rightarrow u$~%;~$u_{emb} = \mathbf{Z}[u_{curr}]$
   \STATE $v_{curr} = m_q \rightarrow v$~%;~$v_{emb} = \mathbf{Z}[v_{curr}]$
   \STATE $u_{surv} = 0.0~;~v_{surv}=0.0$
   \STATE
   \FOR {$N$ samples}
   \STATE select $u_{other} \in \mathbf{l}$ uniformly randomly s.t. $u_{other} \notin \{u_{curr}, v_{curr}\} $
   \STATE select $v_{other} \in \mathbf{l}$ uniformly randomly s.t. $v_{other} \notin \{u_{curr}, v_{curr}\} $
   \FOR {$k \in \{0, 1\}$}
   \STATE $u_{surv} += \lambda_k^{u_{curr},v_{other}}(t_{curr})$
   \STATE $v_{surv} += \lambda_k^{u_{other},v_{curr}}(t_{curr})$
   \ENDFOR
   \ENDFOR
   \STATE$L_{surv} += (u_{surv} + v_{surv}) / N$
   \ENDFOR
   \STATE \textbf{return} $L_{surv}$
\end{algorithmic}
\end{algorithm}

\vspace{-3mm}
\section{Applications}
\vspace{-1mm}
Here, we show that our framework unifies various applications. Specifically we model dynamical processes over graph structured data using the DyRep framework and use the intensity based inference mechanism to predict both structural and behavioral evolution over time in the form of predicting association/communication links. We also support time prediction for the next event to occur between any two nodes. 
\vspace{-2mm}
\subsection{Dynamic Link Prediction}
\vspace{-2mm}
When any two nodes in a graph has increased rate of interaction events, they are more likely to get involved in further interactions and eventually these interactions may lead to the formation of structural link between them. Similarly, formation of the structural link may lead to increased likelihood of interactions between newly connected nodes. To understand, how well our model captures these phenomenon, we ask questions like: 1.) Which is the most likely node $u$ that would undergo an event of type $k$ with a given node $v$ at time $t$? or 2.) Given two nodes $u$ and $v$ at time $t$, which is most likely event type they will participate in?
Its conditional density at time $t$ can be computed:
\begin{equation}\label{density}
\vspace{-2mm}
f^{u,v}_k(t) = \lambda^{u,v}_k(t) \cdot \exp\left(\int_{\bar{t}}^t\lambda(s)ds\right)
\end{equation}
where $\bar{t}$ is the time of the most recent event on either dimension $u$ or $v$. We directly use this conditional density to make most likely node and event type predictions.
\vspace{-2mm}
\subsection{Event Time Prediction}
This is a relatively novel application where the aim is to compute the next time point when a particular type of event (structural or interaction) can occur. Given a pair of nodes $(u, v)$ and event type $k$ at time $t$, we use Eq.~\ref{density} to compute conditional density at time $t$. The next time point $\hat{t}$ for the event can then be computed as:
$
\hat{t} = \int_{t}^\infty tf^{u,v}_k(t)dt
$
where the integral does not have an analytic form and hence we estimate it using Monte Carlo trick.

\vspace{-2mm}
\section{Experiments}
\vspace{-2mm}
\subsection{Datasets}
\vspace{-2mm}
We evaluate DyRep and baselines on two real world datasets: Social Evolution Dataset released by MIT Human Dynamics Lab and Github Dataset available at Github Archive. Table~\ref{tab:tab_data} provides statistics for our final dataset used in the experiments. These datasets cover a range of configurations as Social Dataset is a small network with high clustering coefficient and over 2M events. In contrast, Github dataset forms a large network with low clustering coefficient and sparse events thus allowing us to test the robustness of our model.

\begin{table}[h!]
\label{tab_data}
\caption{Dataset Statistics for Social Evolution and Github.}
\resizebox{1\textwidth}{!}{
\begin{tabular}{ccccccc}
\toprule
Dataset & \#Nodes & \#Initial & \#Final & \#Communications & Clustering\\
 &  & Associations & Associations & &Coefficient&\\
\midrule
Social Evolution & 100 & 407 & 809 & 2020554 & 0.548\\
Github & 12328 & 70640 & 166565 & 604954 &  0.087\\
\bottomrule
\end{tabular}
}
\label{tab:tab_data}
\end{table}

\subsection{Baselines}
For Dynamic Link Prediction task, we compare the performance of our model against multiple representation learning  baselines, three of which has capability to model evolving graphs. Specifically, we compare with Know-Evolve~\cite{TriDaiWanSon17}, DynGem~\cite{GoyKamHeLiu17}, GraphSage~\cite{HamYinLes17} and Node2Vec~\cite{GroLes16}. Table~\ref{tab:compare} provides qualitative comparisons between all methods. Below we describe each of them in detail:

\begin{itemize}

\item Know-Evolve~\cite{TriDaiWanSon17}: This work is the state-of-art model for multi-relational dynamic graphs where each edge has time-stamp and type (communication events).  It models the occurrence of an edge as a multivariate point process whose intensity
function is modulated by the score for that
edge computed based on the learned entity embeddings. The temporally evolving entity embeddings are learned via recurrent architecture. They do not have time-scale specific parameter in intensity function and also does not model graph structure or association links. 
\item DynGem~\cite{GoyKamHeLiu17}: It divides timeline into discrete time points and learns embedding for the graph snapshots at these time points. Specifically it employs autoencoder model and learns embedding in a warm start manner by using the learned embeddings from previous snapshot to initialize the training of current snapshot. To support growing nodes, they propose a  heuristic quantity PropSize, to dynamically determine the number of hidden units required
for each snapshot. They do not model time explicitly.
\item GraphSage~\cite{HamYinLes17}: While not explicitly designed for dynamic or temporally evolving graphs, this is an inductive representation learning method that learns sample and aggregation functions to learn representations instead of training for individual node. This makes it powerful baseline for modeling association links in our setup as they inherently support new nodes and edges due to its inductive capability. 
\item Node2Vec~\cite{GroLes16}: This is a simple baseline to learn graph embeddings over static graphs with a smart random walk based approach to select the neighborhood to learn individual node embeddings. We compare with this baseline to include a purely static and transductive baseline as part of our evaluation.
\end{itemize}

For Event Time Prediction, we compare our model against Know-Evolve described above which also has the capability to predict time in a multi-relational dynamic graphs. Further, we compare with purely temporal model where all events in graph are considered as dyadic and modeled as Multi-dimensional Hawkes Process (MHP) ~\cite{DuWanHeetal15}. 

\begin{table}[t!]
\caption{Comparison of DyRep with state-of-art approaches}
%\footnotesize
\resizebox{1\textwidth}{!}{
\begin{tabular}{ccccc}
\toprule
Key & DyRep & Know-Evolve & DynGem & Static Methods\\
Properties & (Our Method) & & & (e.g. GraphSage) \\
\midrule
Models Association & \checkmark & X & \checkmark & \checkmark\\
Models Communication & \checkmark & \checkmark & X & X \\
Models Time & \checkmark & \checkmark & X & X \\
Learns Representation & \checkmark & \checkmark & \checkmark & \checkmark \\
Graph Information & Attended Non-Uniform & Single & Complete & Uniformly Sampled\\
& 2nd-order & Edge & 1st and 2nd-order & 2nd-order\\
& Neighborhood &  & Neighborhood &  Neighborhood\\
Predicts Time & \checkmark & \checkmark & X & X\\
\bottomrule
\end{tabular}
}
\label{tab:compare}
\vspace{-3mm}
\end{table}

\subsection{Evaluation Scheme}
We divide our test sets into $n (=6)$ slots based on time and report the performance for each time slot, thus providing comprehensive temporal evaluation of different methods. This method of reporting is expected to provide fine-grained insights on how various methods perform over time as they move farther from the learned training history. 

For the baselines that do not explicitly model time (DynGem, GraphSage and Node2Vec), we adopt a sliding window training approach with warm-start method where we learn on initial train set and test for the first slot. Then we add the data from first slot in the train set and remove equal amount of data from start of train set and retrain the model using the embeddings from previous train. This will give the baselines an opportunity to learn from test data as our model also updates the node embeddings during test (but freeze model parameters after training). 
\\\\ 
{\bf Dynamic Link Prediction.} For a given test record $(u, v, t, l, k)$, we replace $v$ with other entities in the graph and compute the density in~(\ref{density}). We then rank all the entities in descending order of the density and report the rank of the ground truth entity. Please note that the latest embeddings of the nodes update even during the test while the parameters of the model remaining fixed. Hence, when ranking the entities, we remove any entities that creates a pair already seen in the test.
We report Mean Average Rank and HITS(@10) metric for dynamic link prediction.\\\\ 
{\bf Time Prediction.} For a given test record $(u, v, t, l, k)$, we report the next time this communication event may occur. We compute the next time point using the previous scheme and report MAE against the ground truth.
%{\bf Node Popularity Prediction.} We computed groud truth node popularity using~(\ref{popularity}) for all nodes at six different time points corresponding to our reporting outline. For prediction, we use~(\ref{popularity_hat}) to get the popularity values of all nodes at those time points. We the nrank all nodes based on theeir popularity in both cases. At each of the six time points, we report Kendall-$\tau$ correlation between the ranked list of nodes.    

\subsection{Predictive Results}
\begin{figure*}[t!]
\captionsetup{justification=justified}
\small
\centering
\resizebox{1\textwidth}{!}{
\begin{tabular}{cccc}
\includegraphics[width = 0.27\textwidth]{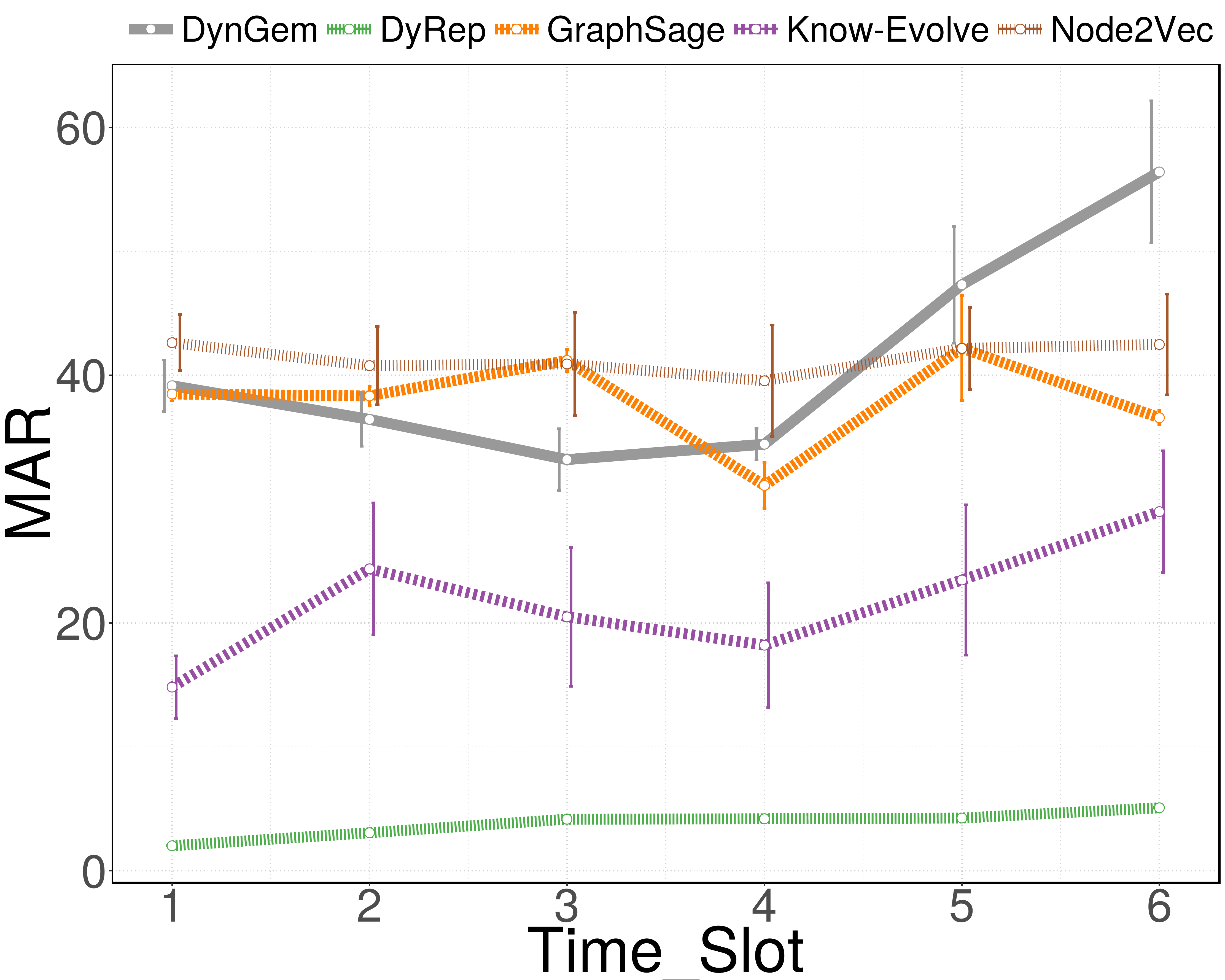}
& \includegraphics[width = 0.27\textwidth]{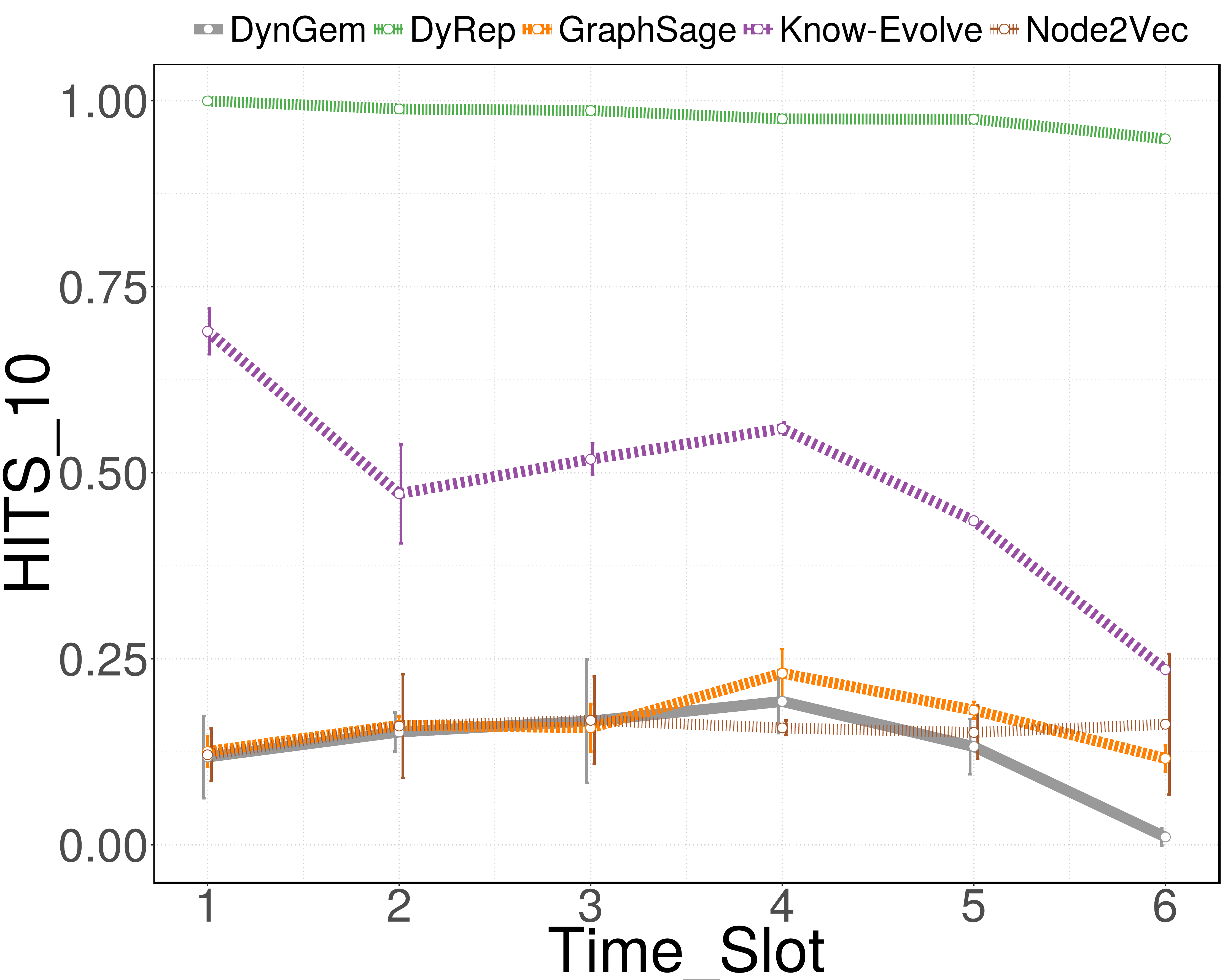}
& \includegraphics[width = 0.27\textwidth]{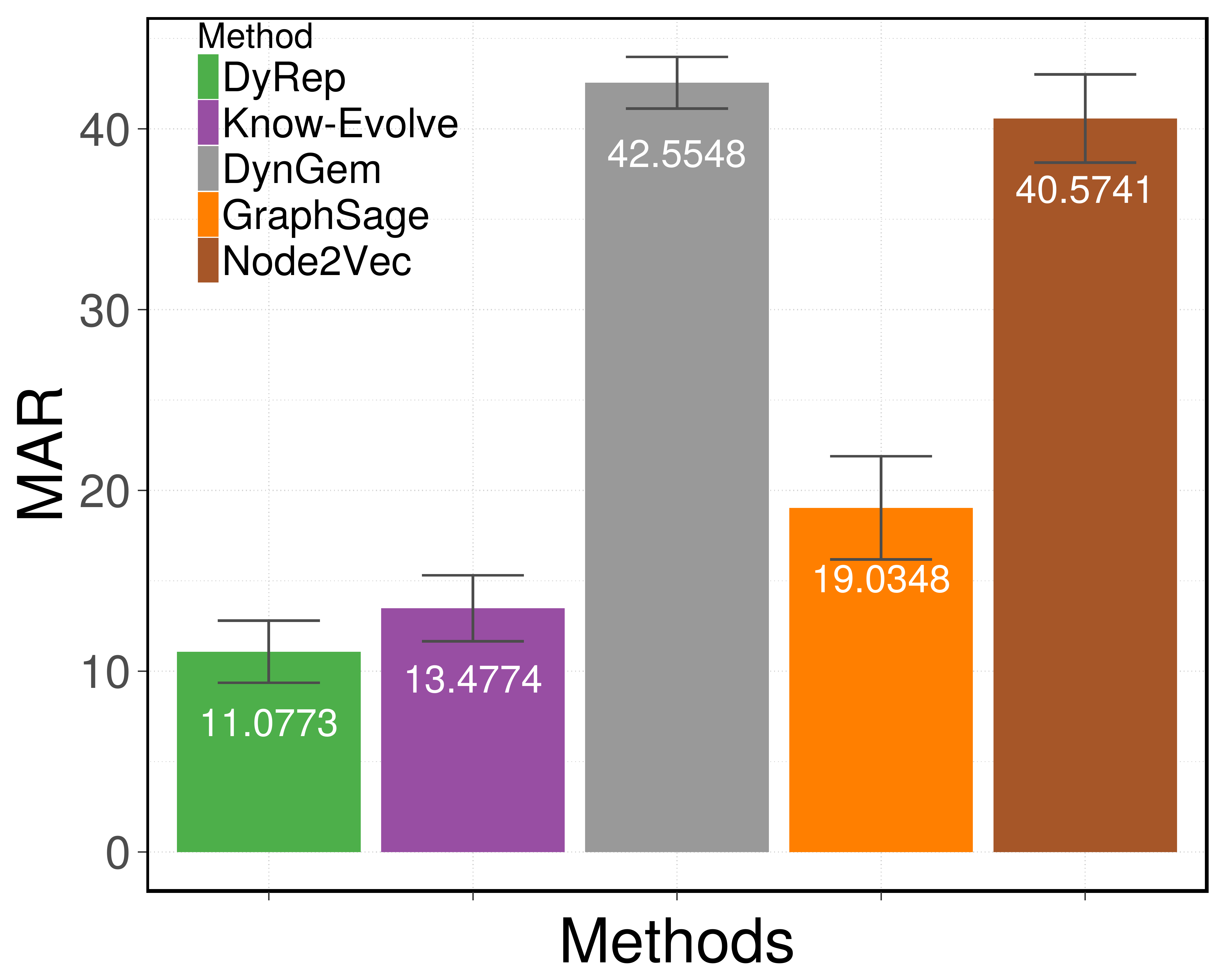}
& \includegraphics[width = 0.27\textwidth]{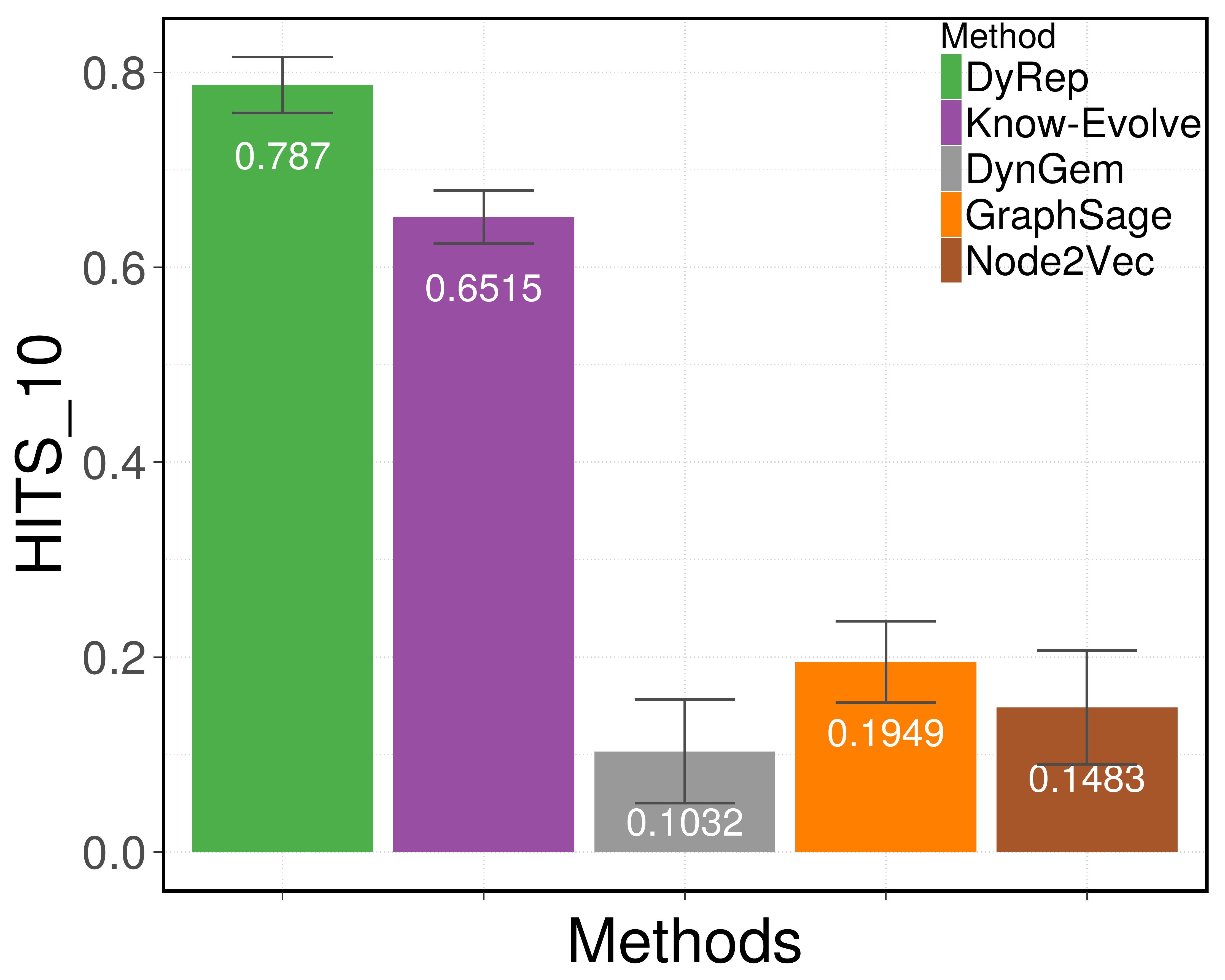}\\
\includegraphics[width = 0.27\textwidth]{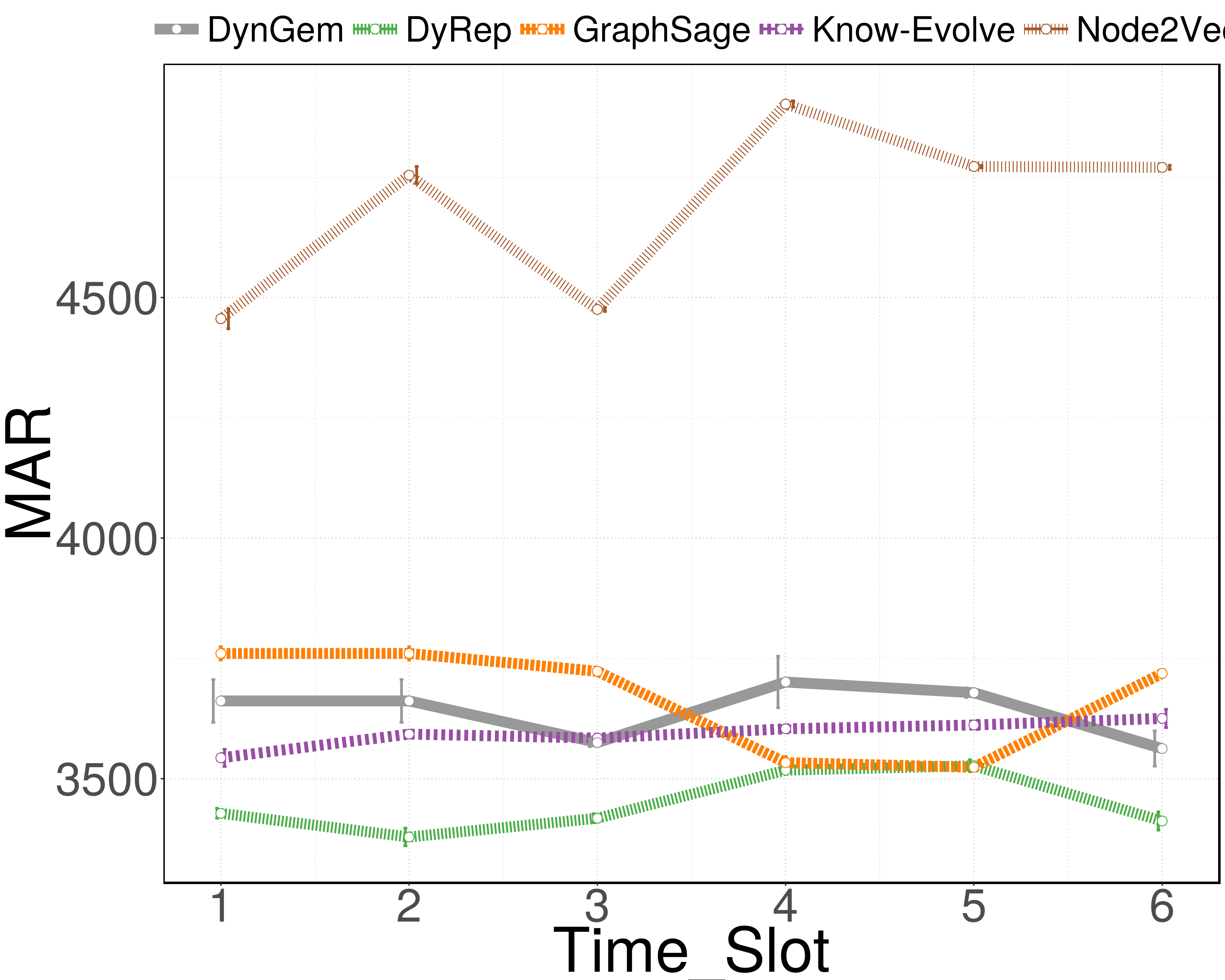}
& \includegraphics[width = 0.27\textwidth]{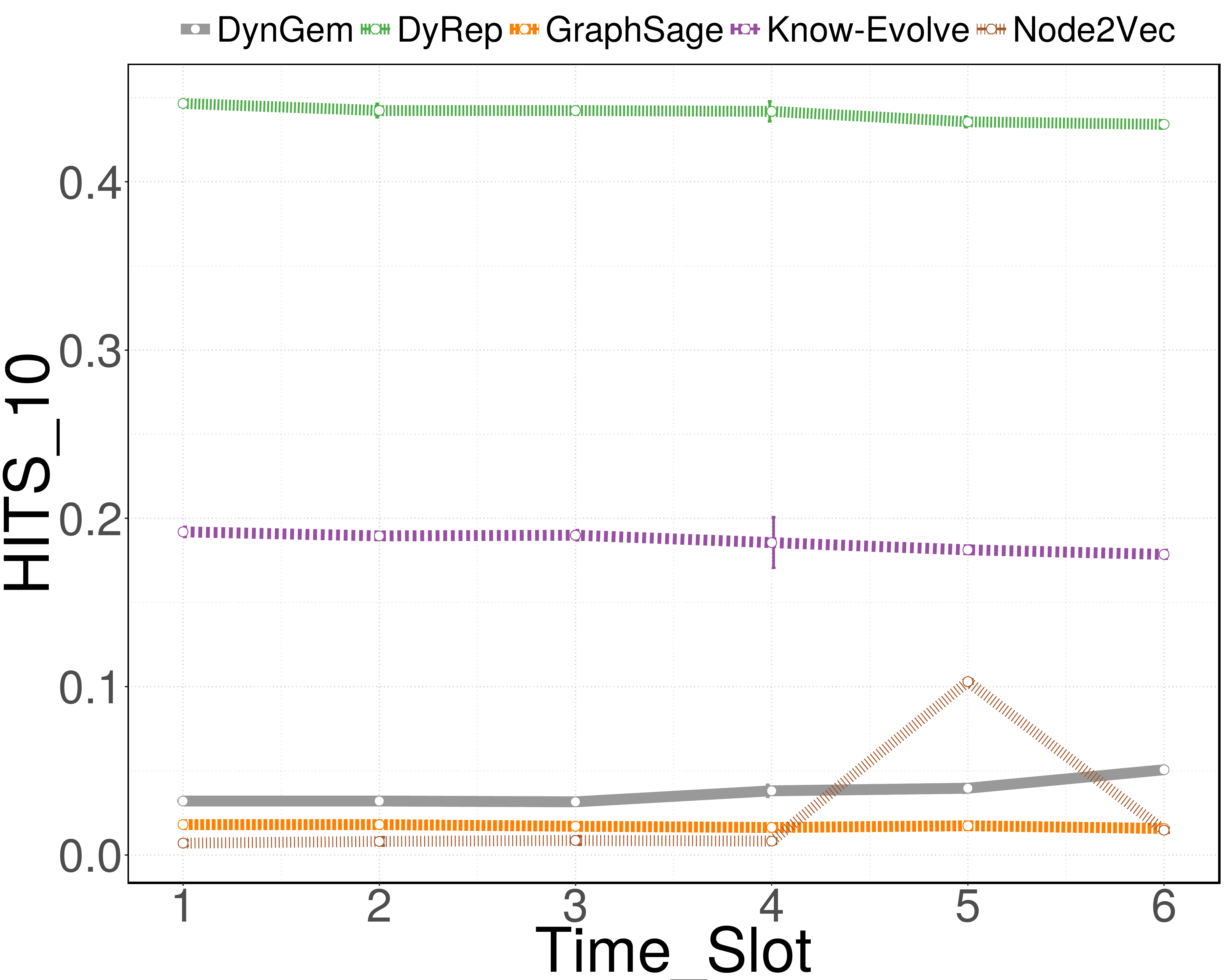}
& \includegraphics[width = 0.27\textwidth]{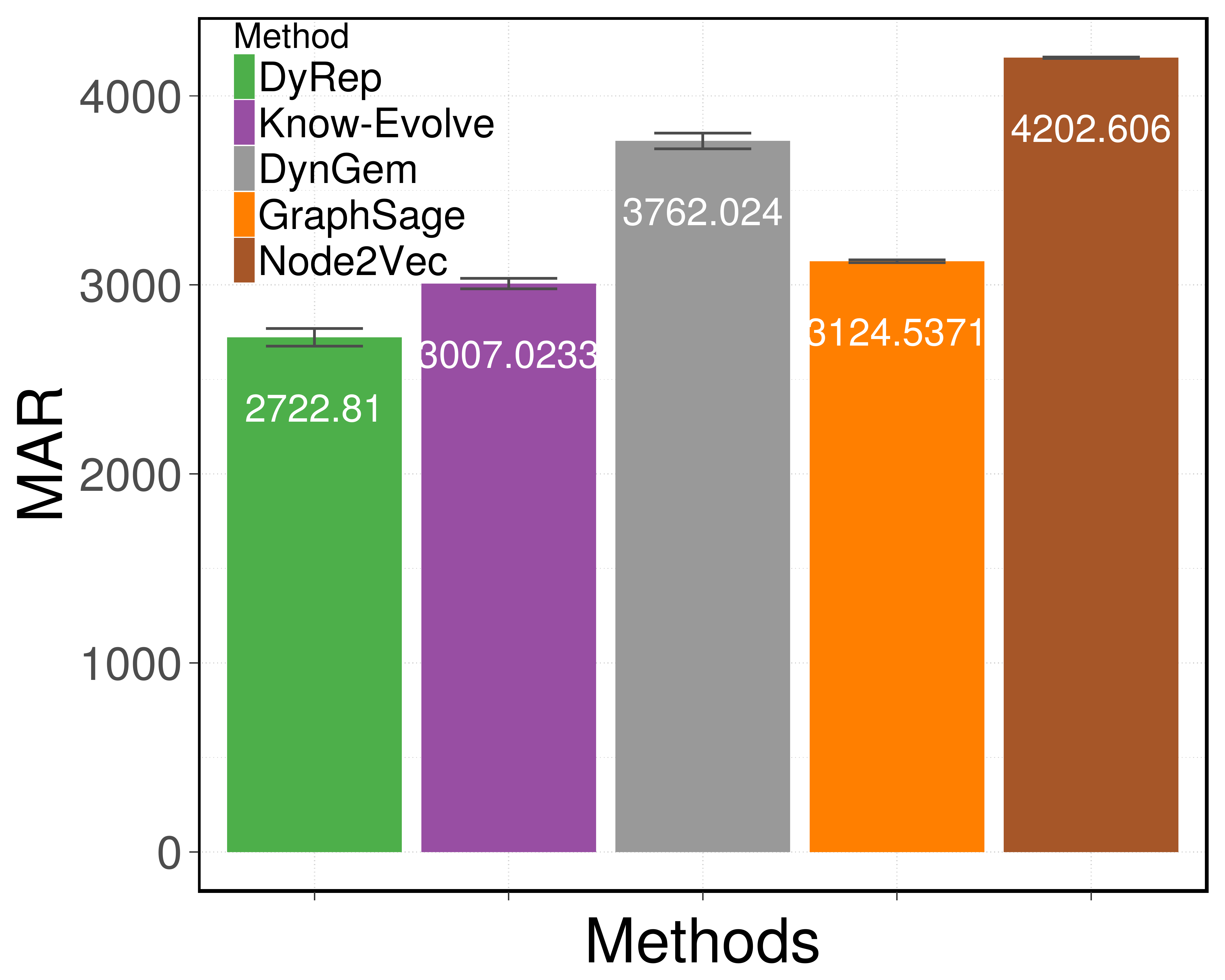}
& \includegraphics[width = 0.27\textwidth]{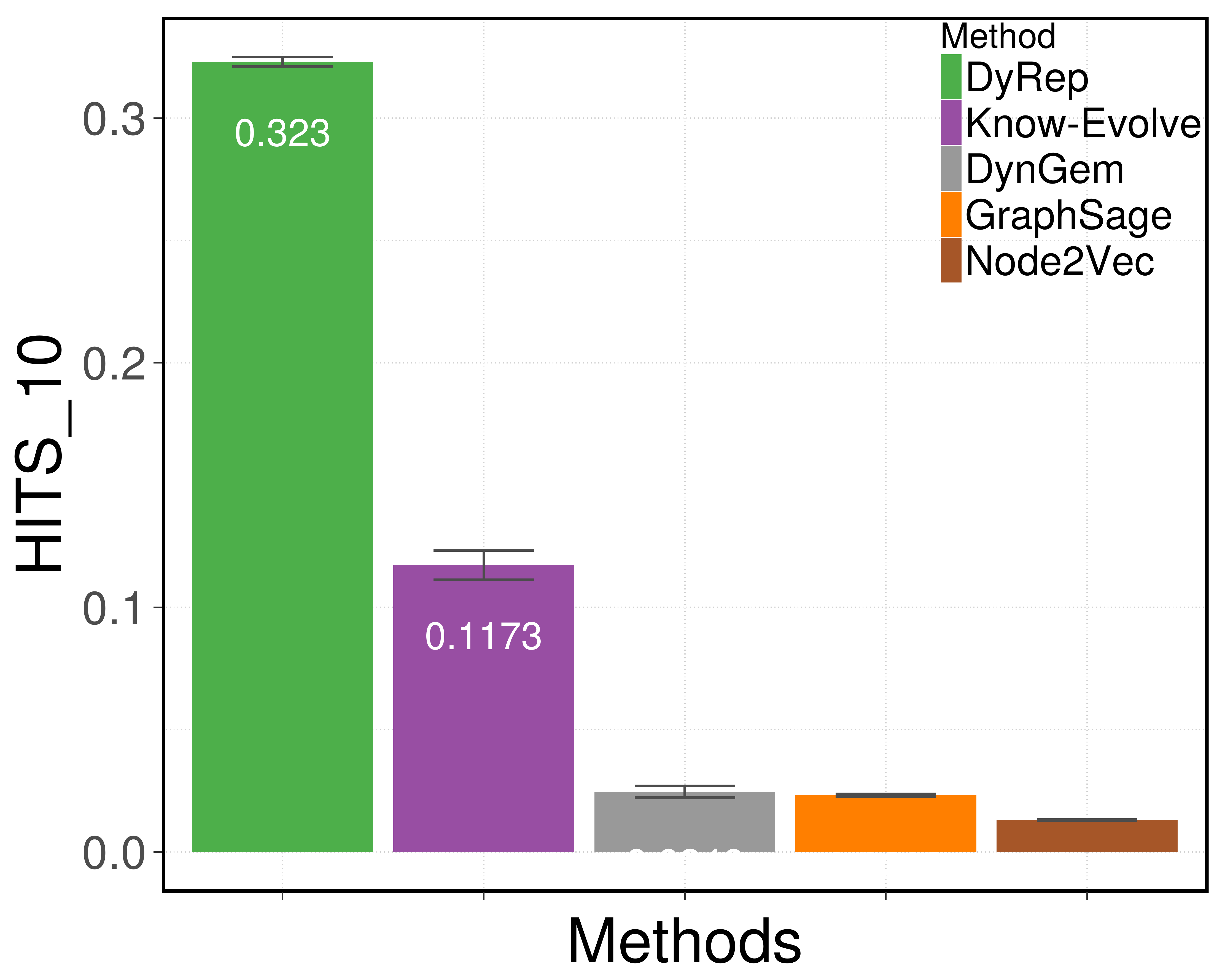}\\
(a) MAR (Communication) & (b) HITS@10 (Communication)&(c) MAR (Association)& (c) HITS@10 (Association)
\end{tabular}
}
\vspace{-3mm}
\caption{Dynamic Link Prediction Performance for {\bf (a-b)} Communication events {\bf (c-d)} Association events. {\bf Top Row:} shows the results for Social Evolution Dataset {\bf Bottom Row:} shows the results for Github dataset}
\label{fig:pred}
\vspace{-3mm}
\end{figure*}

\textbf{Communication Event Prediction Performance.} We first considered the task of predicting communication events between nodes. Please note that such nodes may or may not have a permanent edge (association) between them. The experimental results for this task are reported in columns (a) and (b) of Figure~\ref{fig:pred}.

For Social Evolution dataset, our method  significantly and consistently outperforms all the baselines on both metrics. While the performance of our method drops a little over time, it is expected due to the temporal recency affect on node's evolution. But the overall performance  of our method is very stable across all time points with negligible deviation. Know-Evolve can capture event dynamics well and shows consistently better rank than others but it is interesting that its performance deteriorates significantly in HITS@10 metric over time. We conjecture that  features learned through edge-level modeling limits the predictive capacity of the method over time. Another interesting observation is the inability of DynGem (snapshot based dynamic) and GraphSage (inductive) to significantly outperform Node2vec (transductive static baseline). We believe that discrete time snapshot based models fail to capture fine-grained dynamics of communication events well. Further, while there are many recurrent and new events where time is critical, there is no topological evolution in this graph which seems to amortize GraphSage's inductive ability.

For Github dataset, we demonstrate comparable performance with both Know-Evolve and GraphSage on Rank metric. We would like to note that the overall performance for all methods on rank metric is low. As we reported earlier, Github dataset has very less number of events compared to the number of nodes and it also has very low clustering coefficient which makes it a very challenging dataset to learn. %Further, the sparsity of temporal events for each node makes it more challenging for recurrent network based methods like our model and Know-Evolve baseline. 
It is expected that for a large number of nodes with no communication history, most of the methods will show comparable performance but our method outperforms all others when there is some history available. This is demonstrated by our significantly better performance for HITS@10 metric where we are able to do highly accurate prediction for nodes where we learn better history. This can also be attributed to our model's ability to capture the effect of evolving topology which is missed by Know-Evolve. Finally, we do not see significant decrease in performance of any method over time in this case which can again be attributed to roughly uniform distribution of nodes with no communication history across time slots.

\textbf{Association Event Prediction Performance.} As the occurrence of association events is not available across all time slots, we instead report the aggregate number in columns (c) and (d) of Figure~\ref{fig:pred} for this task. For both the datasets under both metrics, our model significantly outperforms the baselines for this task. Specifically, our model's strong performance on HITS@10 metric across both datasets demonstrates its robustness in accurate learning from various properties of data. On Social evolution dataset, the number of association events are very small (only 485) and hence our strong performance on that dataset shows that the model is able to capture the influence of communication events on the association events through the learned representations. On the Github dataset, there are lots of association events where in many instances the network grows through new nodes. Our model's strong performance across both metric demonstrates its inductive ability to generalize across new nodes across time.

\begin{figure}[t]
\small
\centering
\resizebox{1\textwidth}{!}{
\begin{tabular}{cccc}
\includegraphics[width = 0.25\textwidth]{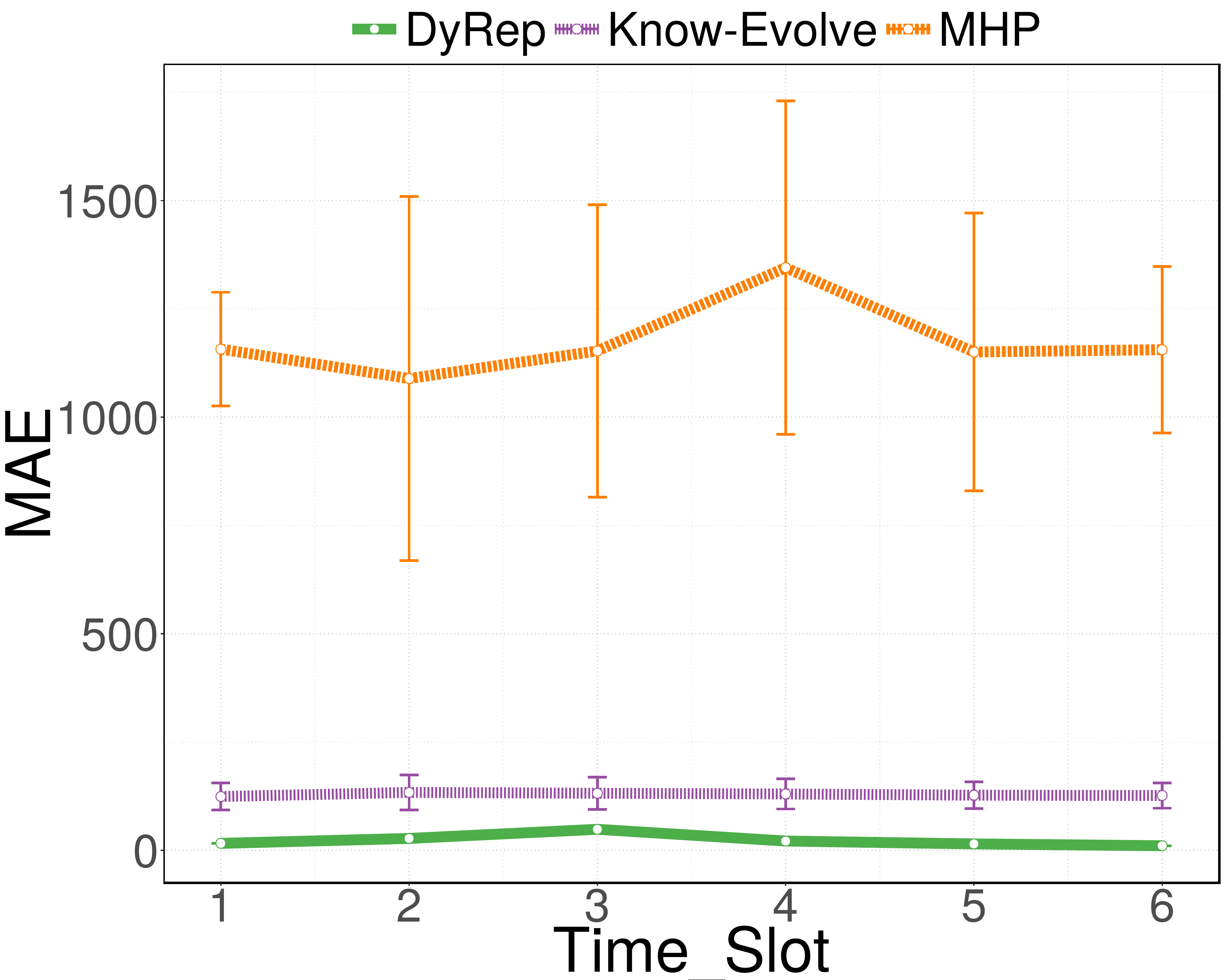}
& \includegraphics[width = 0.25\textwidth]{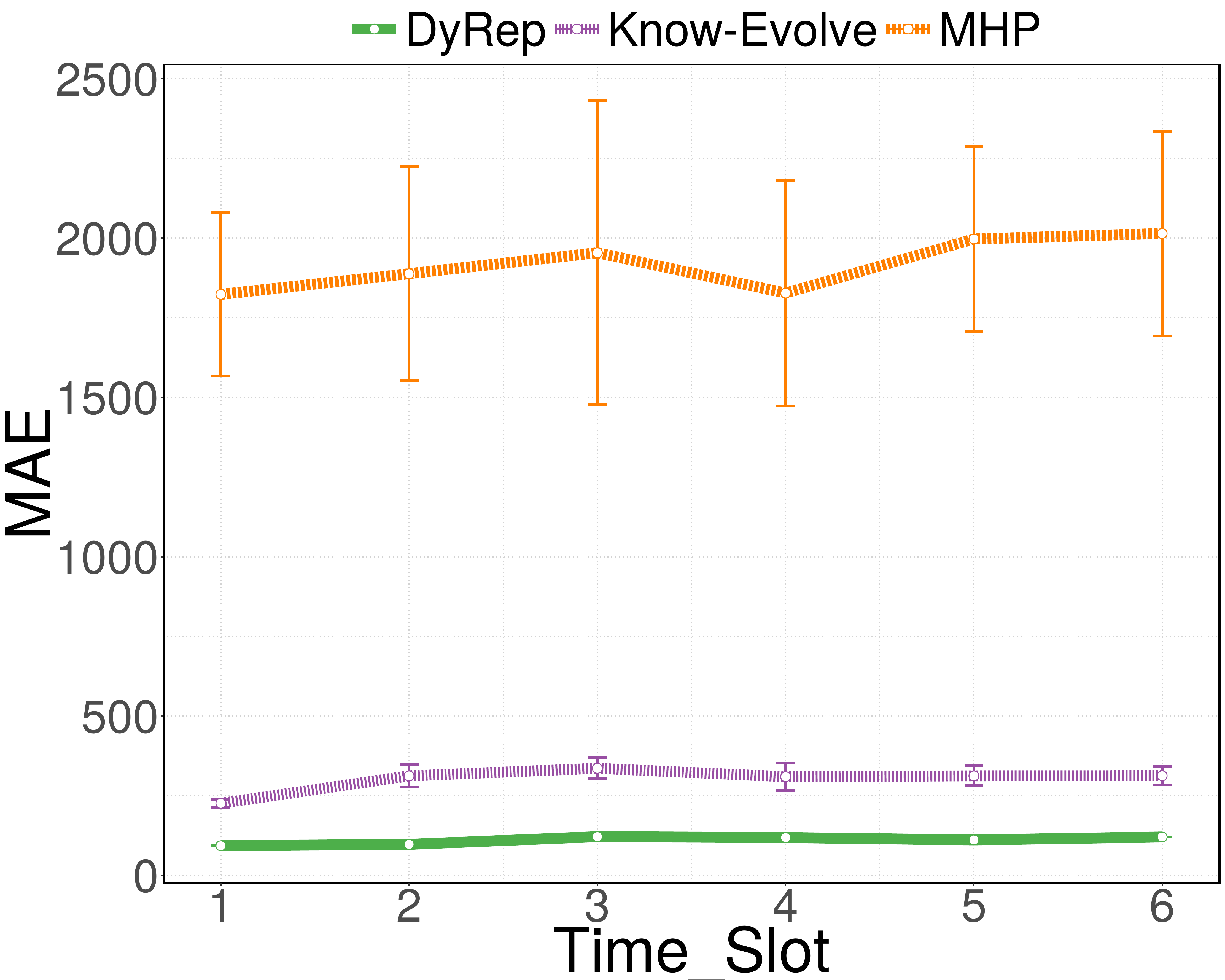}
& \includegraphics[width = 0.25\textwidth]{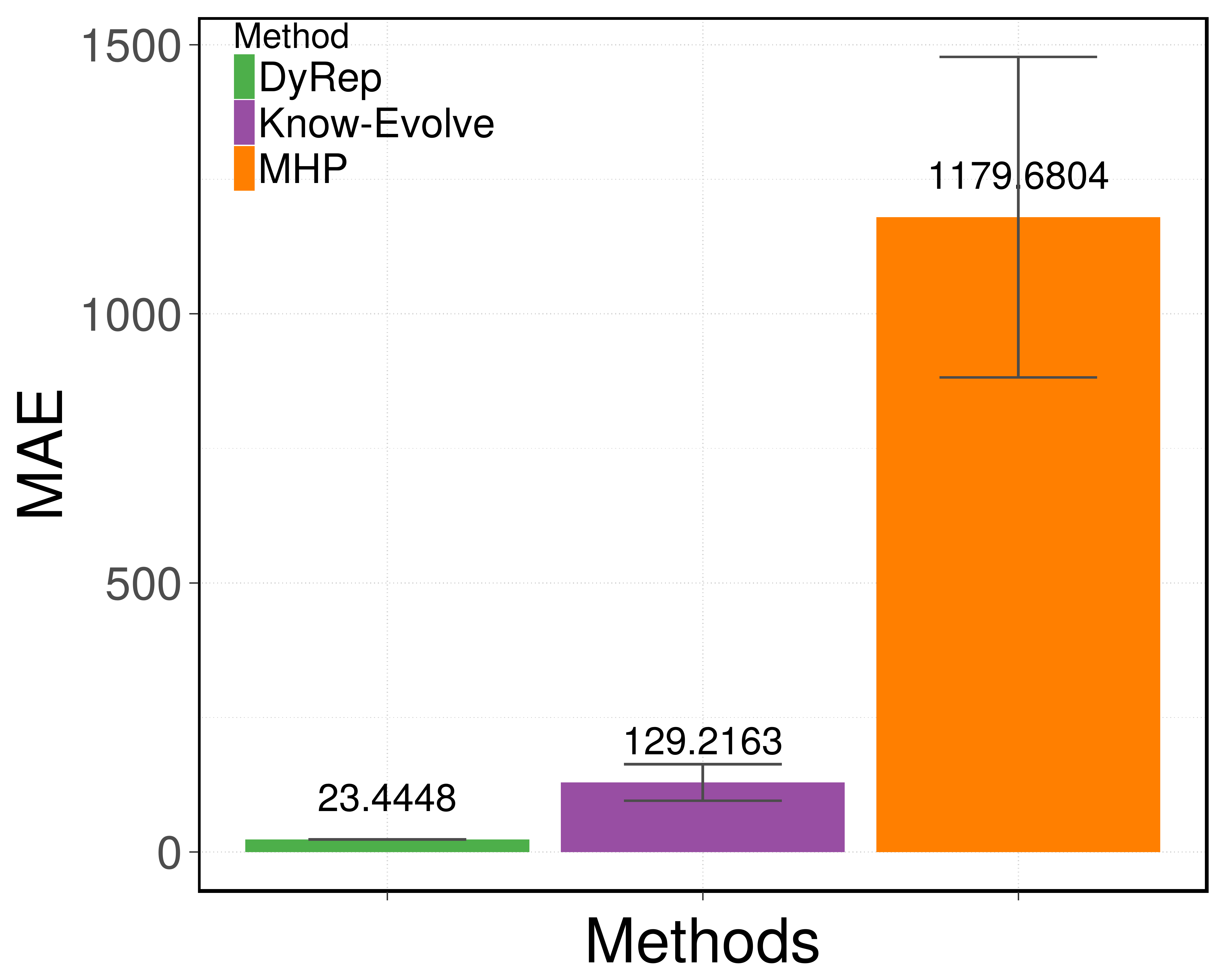}
& \includegraphics[width = 0.25\textwidth]{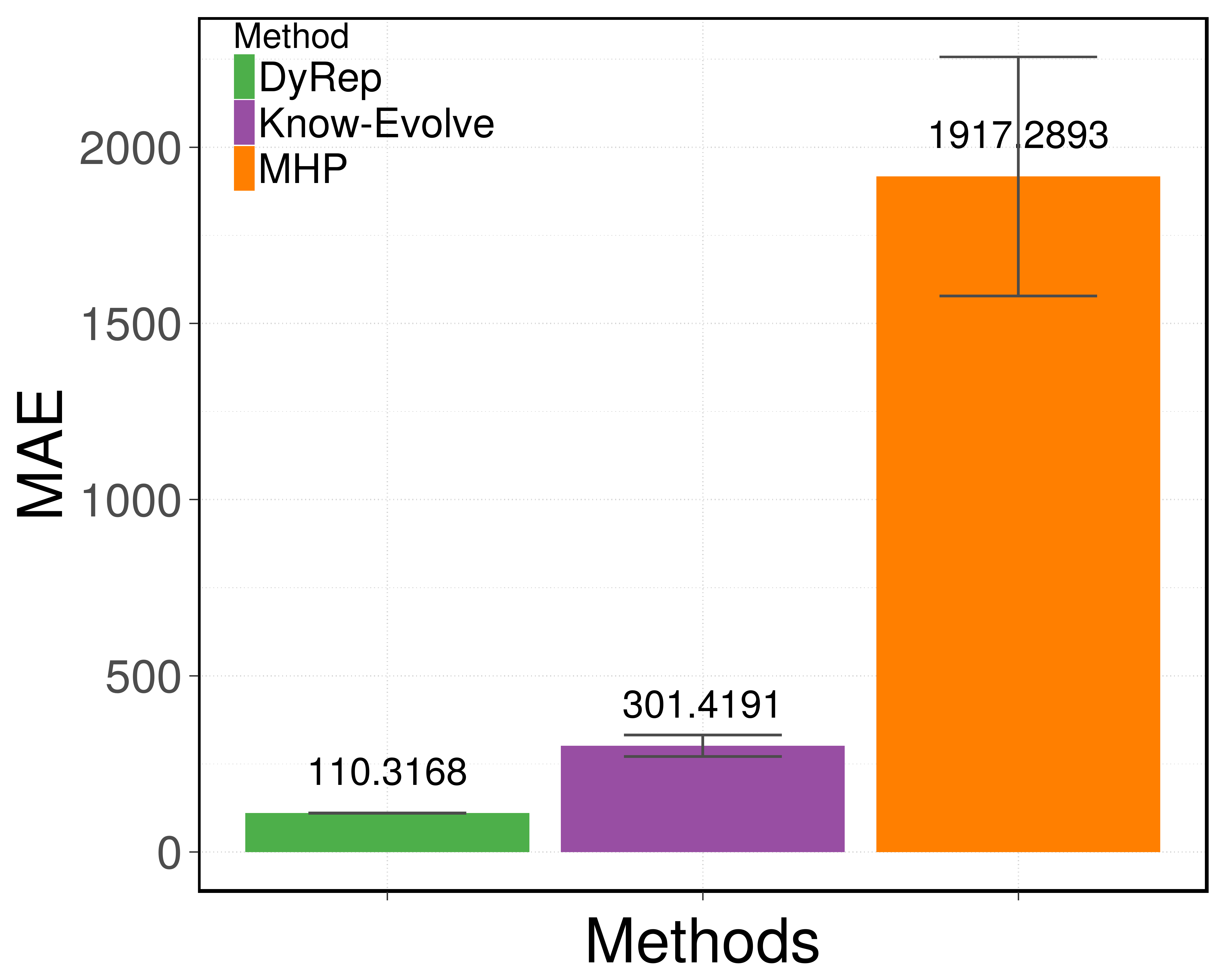}\\
(a) Social Evolution & (b) Github & (c) Social Evolution & (d) Github
\end{tabular}
}
\vspace{-2mm}
\caption{Time Prediction results (unit is hrs).}
\label{fig:time}
\vspace{-3mm}
\end{figure}

\textbf{Time Prediction Performance.} Figure~\ref{fig:time} demonstrates consistently better performance than the state-of-art baseline for event time prediction on both datasets. While Know-Evolve models both processes as two different types of relations between entities, it does not explicitly capture the variance in the time scales of two processes which may explain better performance of our model. Further, Know-Evolve does not consider influence of neighborhood which may lead to capturing weaker temporal dynamics across the graph. MHP uses specific parametric intensity function where each node pair is modeled as an independent dimension which fails to account for intricate dependencies across graph.

%
%\begin{figure}[ht!]
%\small
%\centering
%\begin{tabular}{cc}
%\includegraphics[width = 0.20\textwidth]{kendall_test}
%& \includegraphics[width = 0.20\textwidth]{kendall_test}\\
%(a) Social Evolution & (b) Github
%\end{tabular}
%\vspace{-2mm}
%\caption{Node Popularity Prediction results (Kendall - $\tau$.}
%\label{fig:gdfull}
%\vspace{-2mm}
%\end{figure}
\vspace{-2mm}
\subsection{Qualitative Results}

\begin{figure}[h!]
\small
\centering
\begin{tabular}{cc}
\includegraphics[width = 0.45\textwidth]{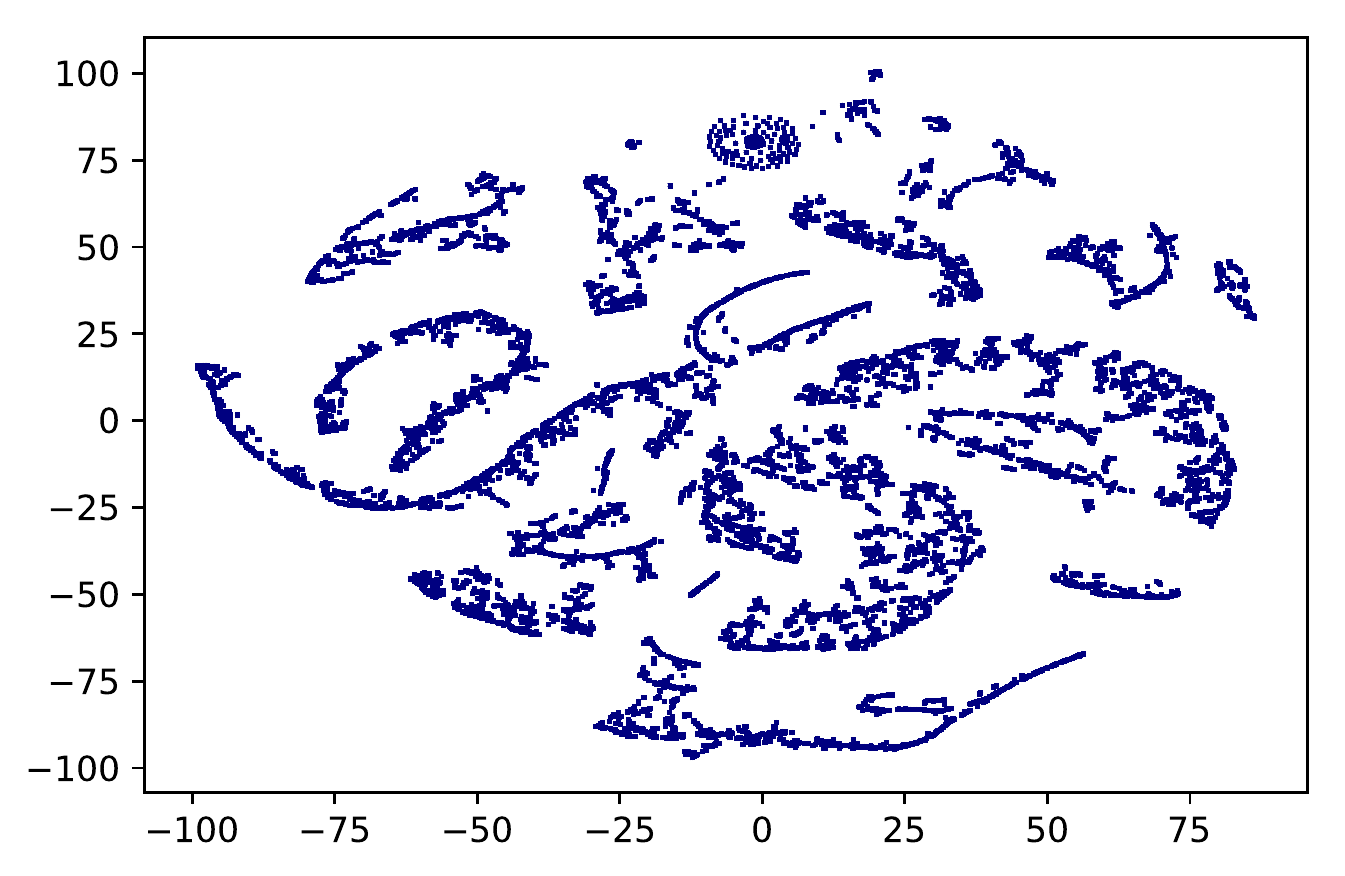}
& \includegraphics[width = 0.45\textwidth]{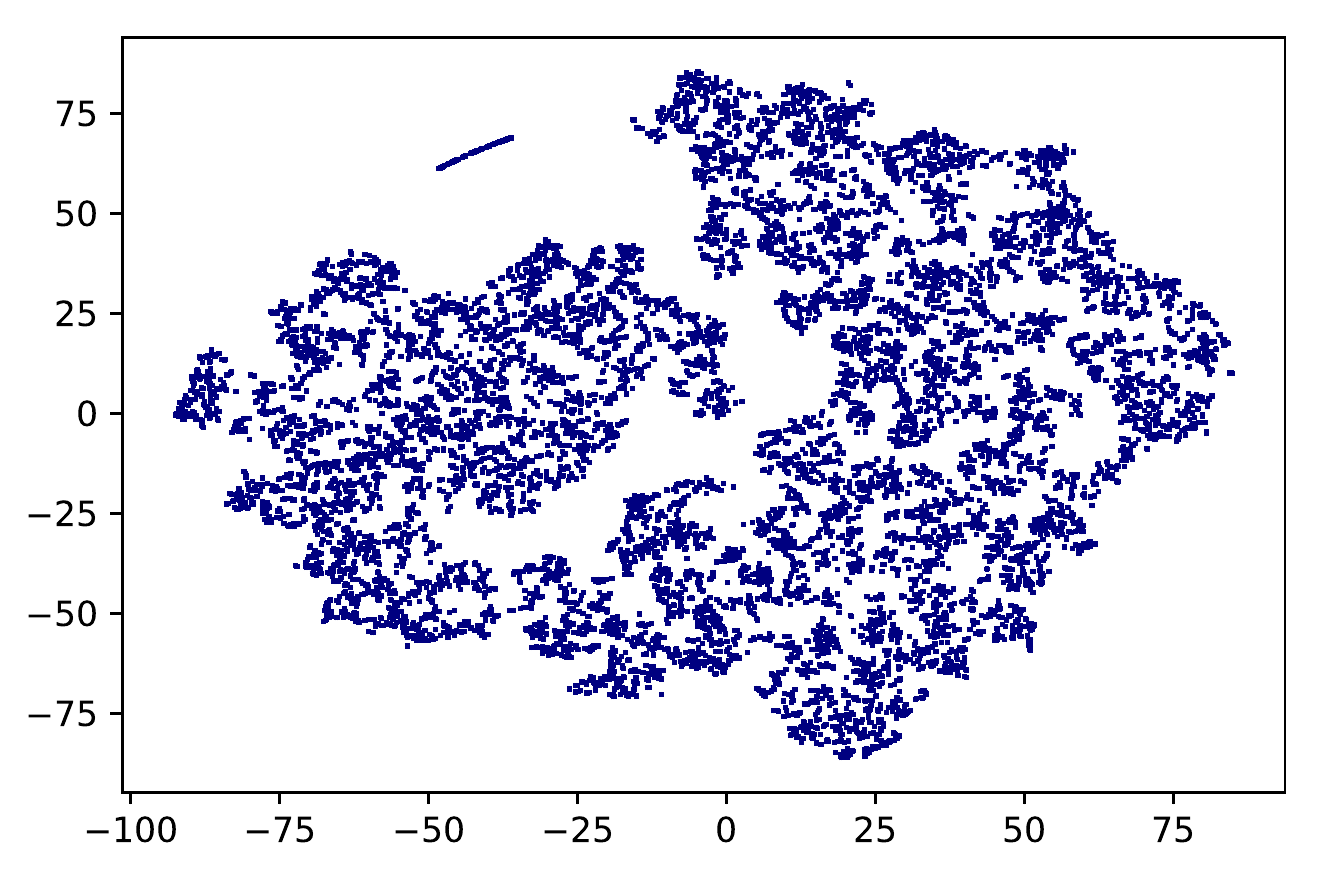}\\
(a) DyRep Embeddings & (b) GraphSage Embeddings 
\end{tabular}
\vspace{-2mm}
\caption{tSNE for learned embeddings after training.}
\label{fig:tsne_embed}
\vspace{-2mm}
\end{figure}

{\bf tSNE Visualization.} We conducted a series of qualitative analysis to understand the discriminative power of evolving embeddings learned by DyRep and see the effect of temporal evolution, association process and communication process on the learned embeddings. For this, we compare our embeddings against the embeddings learned by GraphSage as it is state-of-art static embedding method that is also inductive. Figure~\ref{fig:tsne_embed} shows the tSNE embeddings learned by Dyrep (left) and GraphSage (right) respectively. We used sklearn.manifold.TSNE library to plot this figure with $n\_components=2$, $learning\_rate=200$, $perplexity=30$, $metric="euclidean"$, $min\_grad\_norm=1\mathrm{e}{-9}$, $early\_exaggeration=4$ and ran for 40,000 iterations.
The visualization demonstrates that DyRep embeddings have more discriminative power as it can effectively capture the distinctive and evolving structural features over time as aligned with empirical evidence.

We assess the quality of learned embeddings and the ability of model to capture both temporal and structural information. Let $t_0$ be the time point when train ended. Let $t_1$ be the timepoint when the first test slot ends.

{\bf Effect of Association and Communication on Embeddings.} We conducted this experiment on Social dataset. We consider three use cases to demonstrate how the interactions and associations between the nodes changed their representations and visualize them to realize the effect. 

\begin{itemize}
\item {\bf Nodes that did not have association before test but got linked during first test slot.} Nodes 46 and 76 got associated in test between test points 0 and 1. This reduced the cosine distance in both models but DyRep shows prominent effect of this association which should be the case. DyRep reduces the cosine distance from 1.231 to 0.005. Also, DyRep embeddings for these two points belong to different clusters initially but later converge to same cluster. In GraphSage, the cosine distance reduces from 1.011 to 0.199 and the embeddings still remain in original clusters. Figure~\ref{fig:uc1} shows the visualization of embeddings at the two time points in both the methods. This demonstrates that our embeddings can capture association events effectively.

\begin{figure*}[h!]
%\captionsetup{justification=justified}
\small
\centering
\begin{tabular}{cc}
\includegraphics[width = 0.45\textwidth]{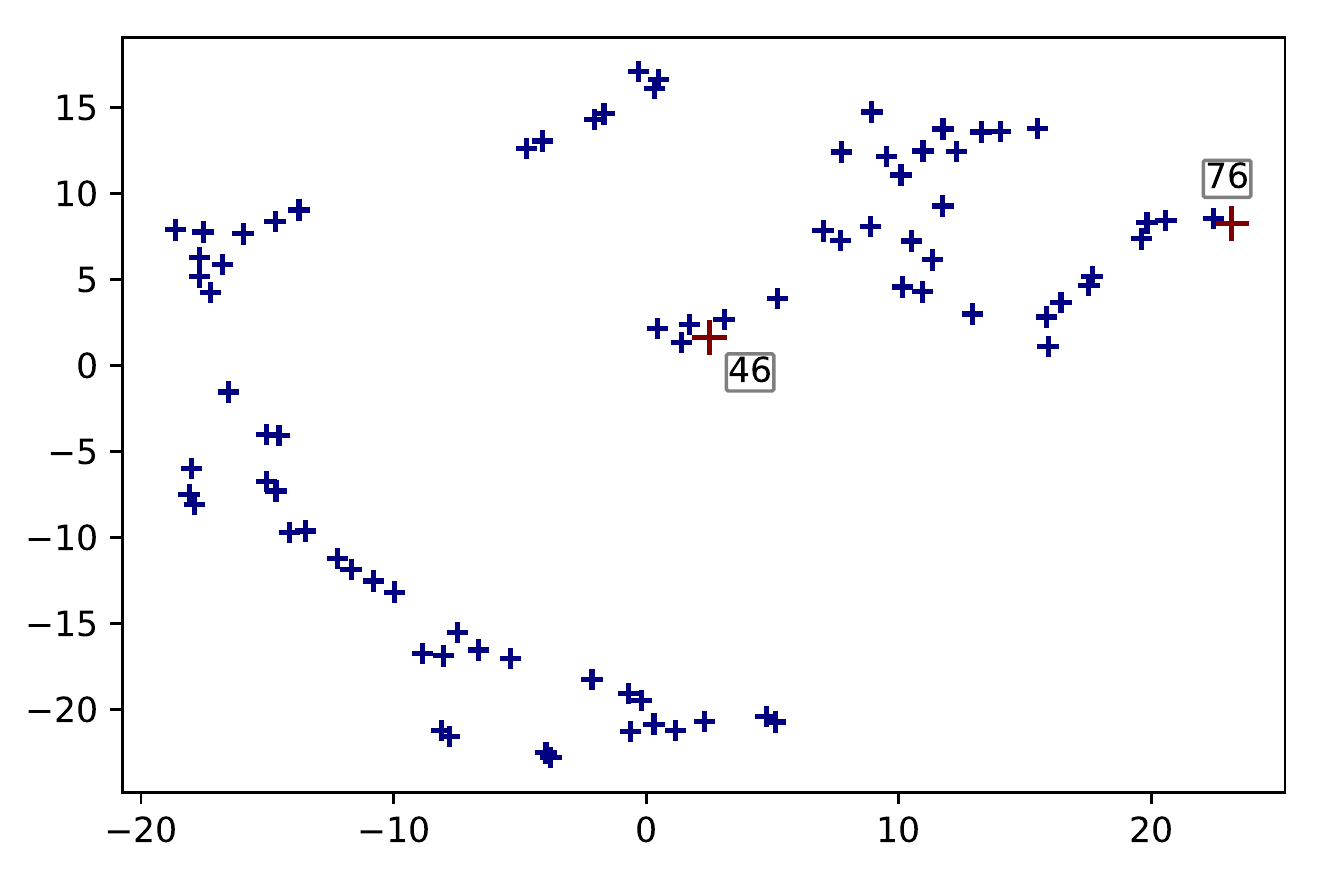}
& \includegraphics[width = 0.45\textwidth]{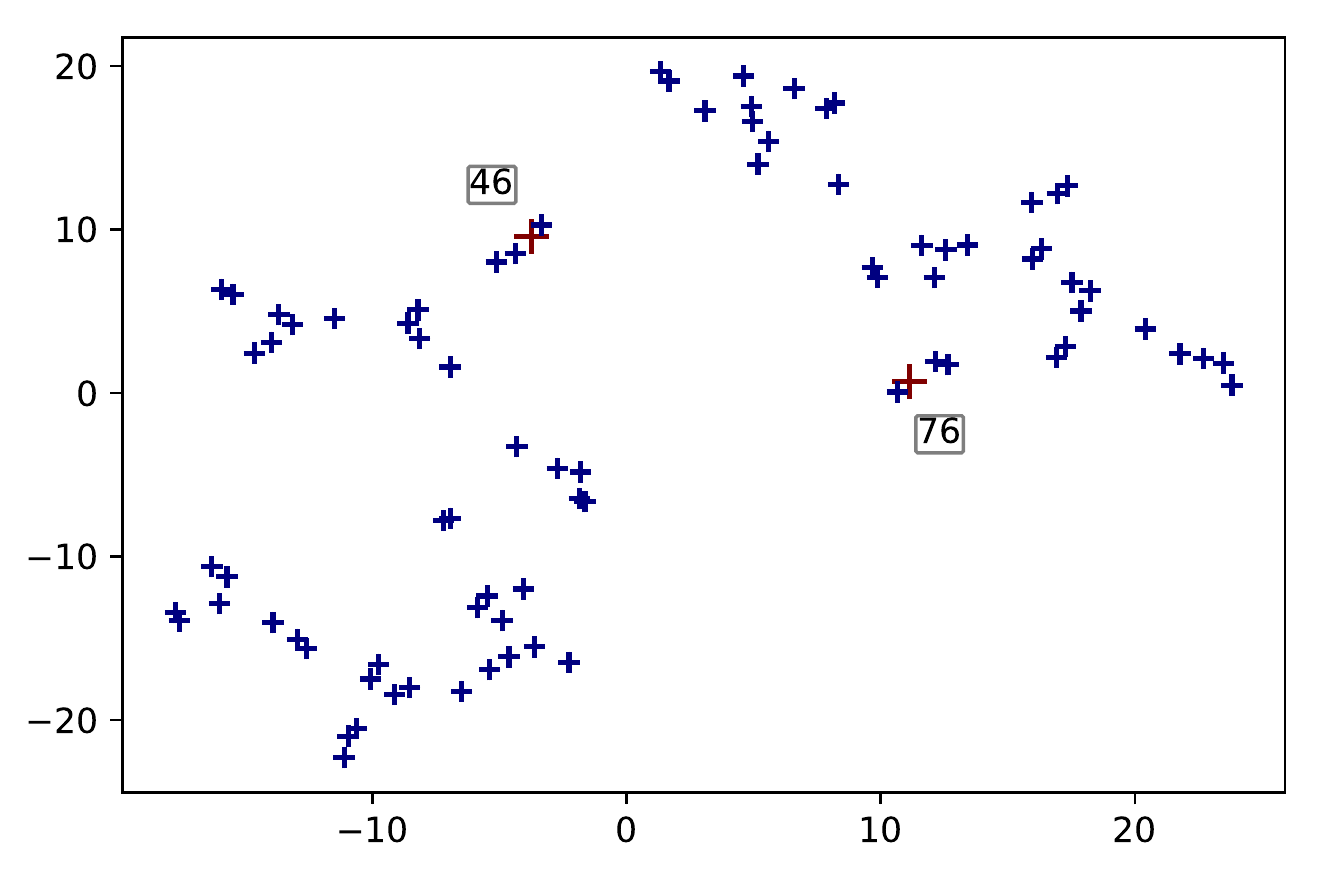}\\
\includegraphics[width = 0.45\textwidth]{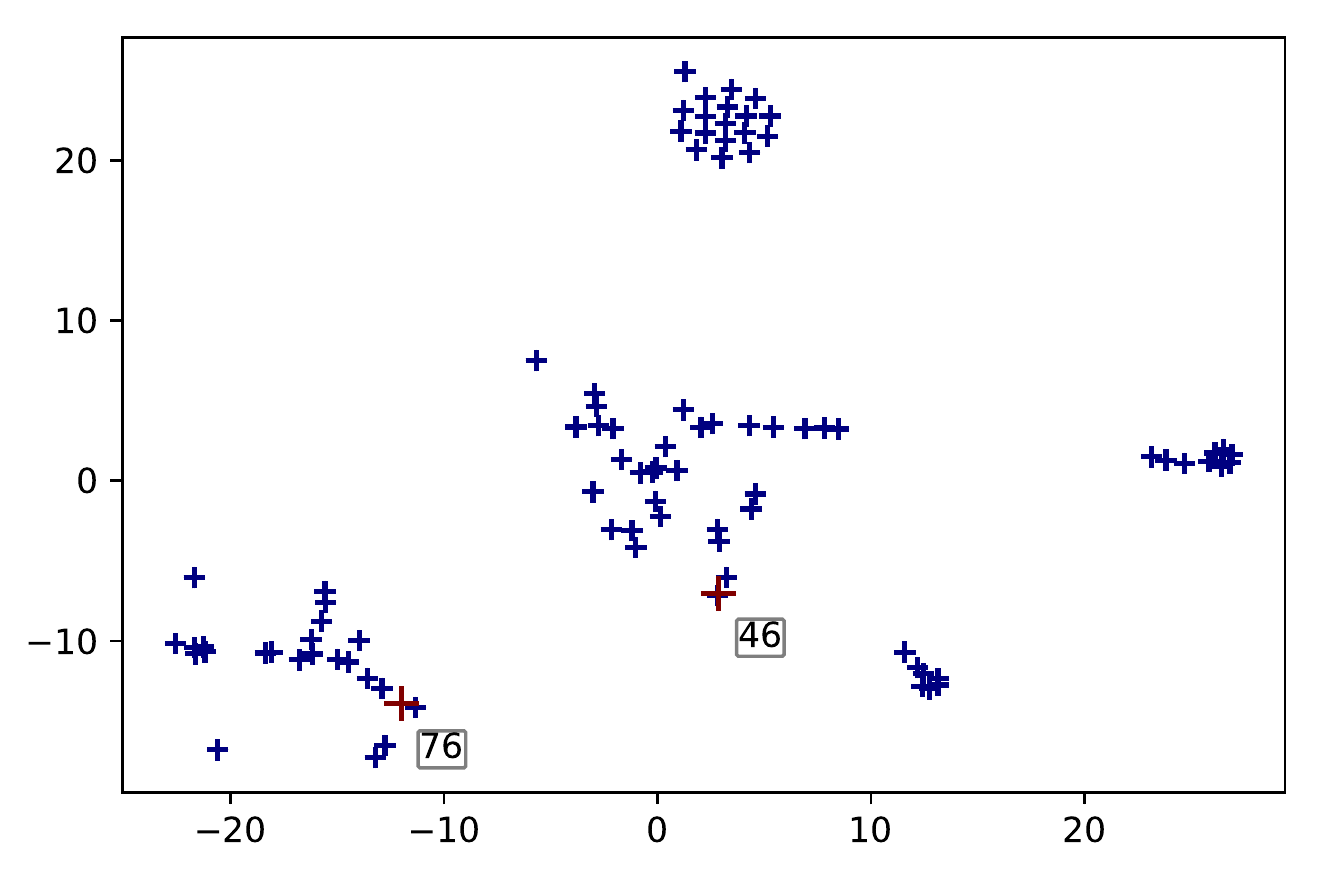}
& \includegraphics[width = 0.45\textwidth]{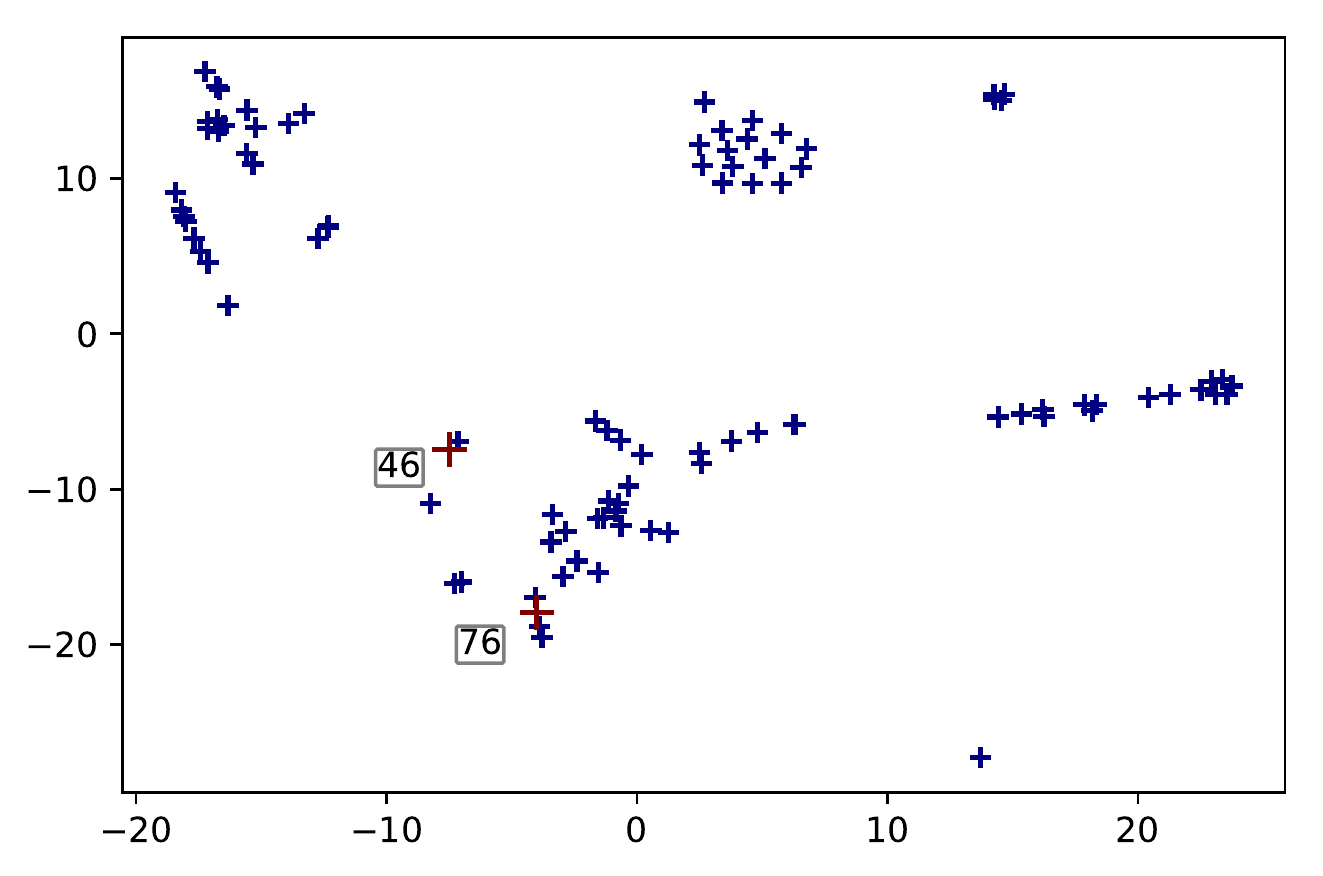}\\
(a) Train End Time & (b) Test Slot 1 End Time. 
\end{tabular}
\vspace{-2mm}
\caption{Use Case I. {\bf Top row:} GraphSage Embeddings. {\bf Bottom Row:} DyRep Embeddings. }
\label{fig:uc1}
\vspace{-2mm}
\end{figure*}

\item {\bf Nodes that did not have association but many communication events (114000).} Nodes 27 and 70 is such a use case. DyRep embeddings consider the nodes to be in top 5 nearest neighbor of each other, in the same cluster and cosine distance of 0.005 which is aligned with the fact that nodes with large number of events tend to develop similar features over time. Graphsage  on the other hand considers them 32nd nearest neighbor, puts them in different clusters with cosine distance - 0.792. Figure~\ref{fig:uc2} shows the visualization of embeddings at the two time points in both the methods. This demonstrates that our embeddings can capture communication events and their temporal effect on embeddings effectively. 

\begin{figure*}[h!]
\small
\centering
\begin{tabular}{cc}
\includegraphics[width = 0.45\textwidth]{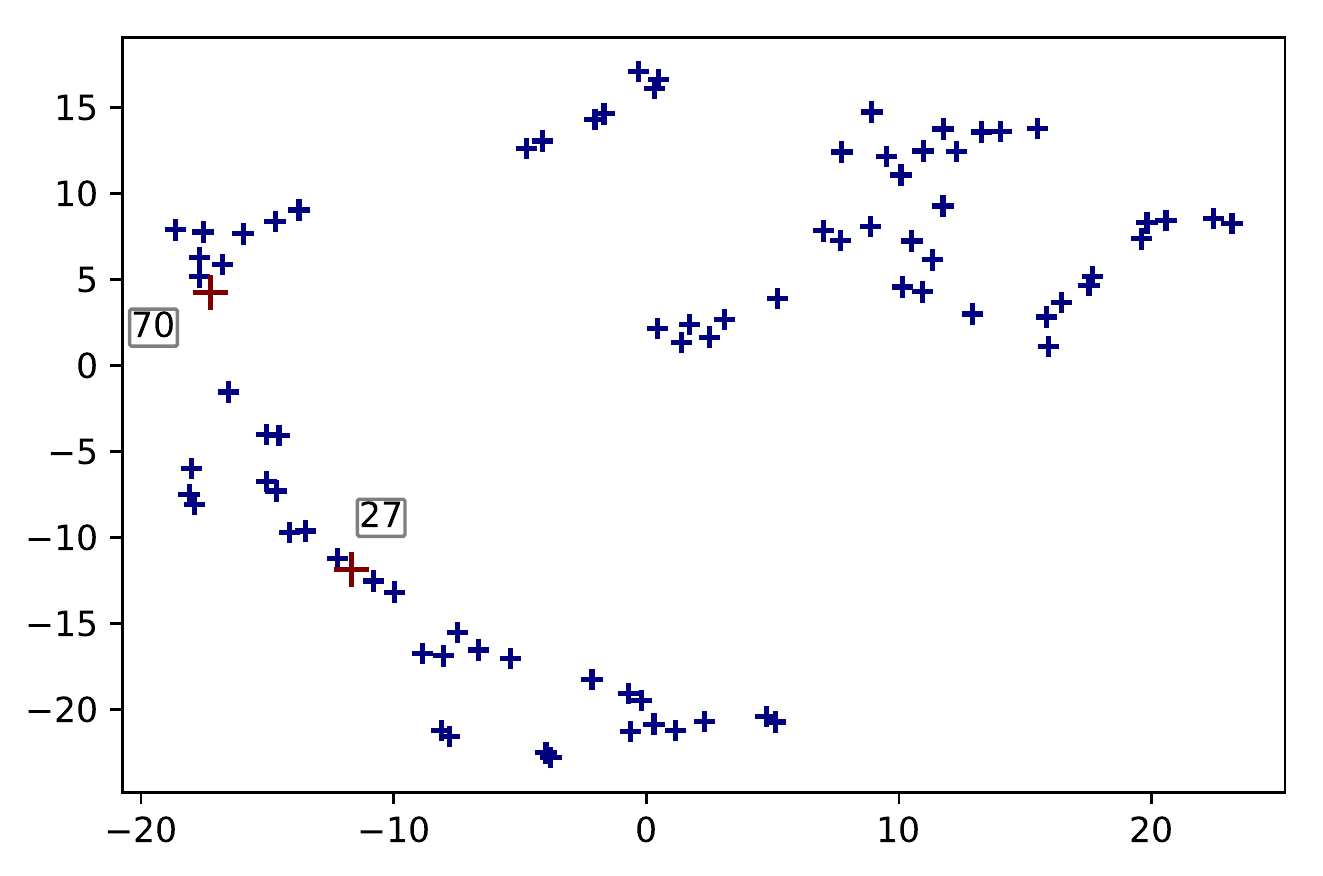}
& \includegraphics[width = 0.45\textwidth]{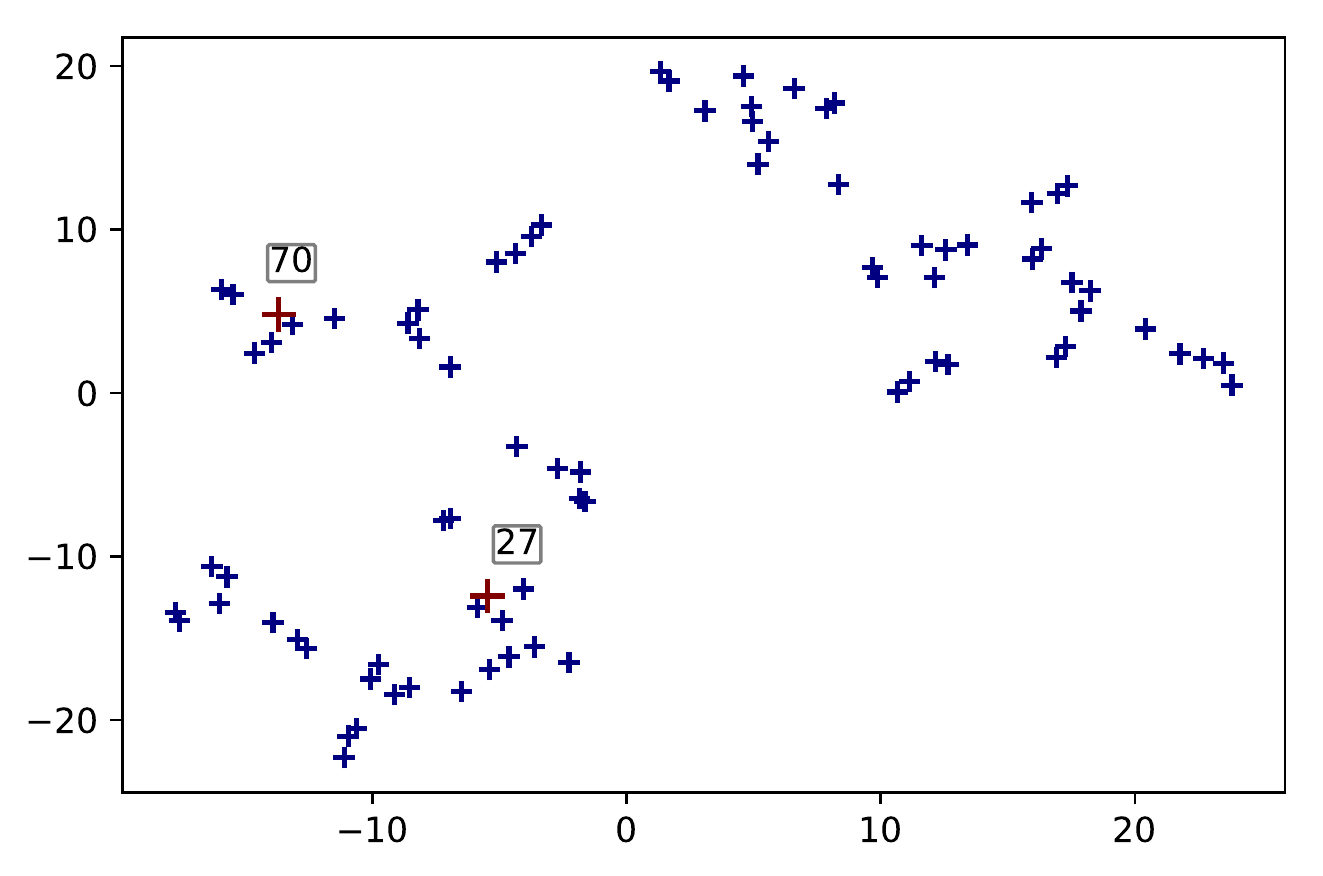}\\
\includegraphics[width = 0.45\textwidth]{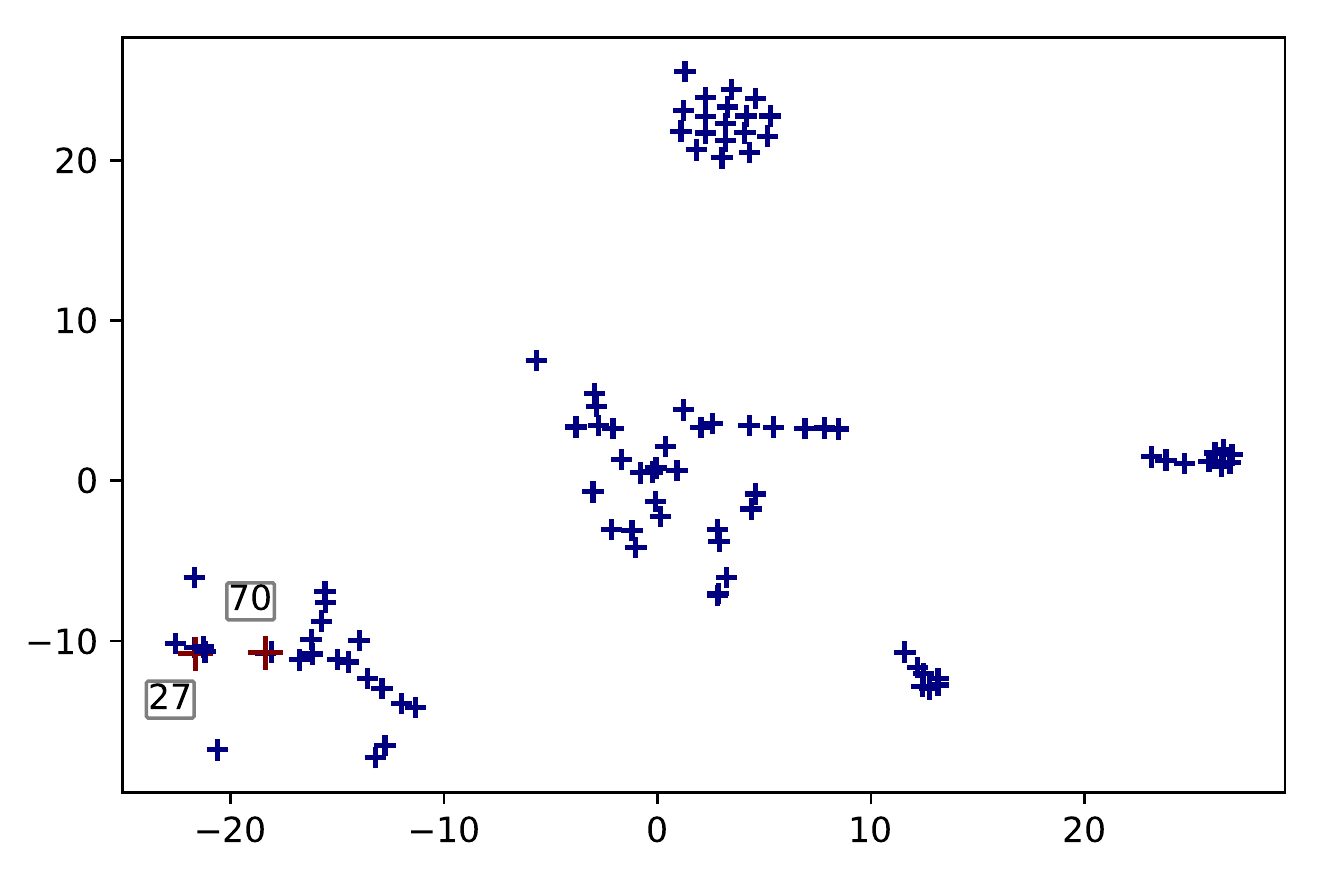}
& \includegraphics[width = 0.45\textwidth]{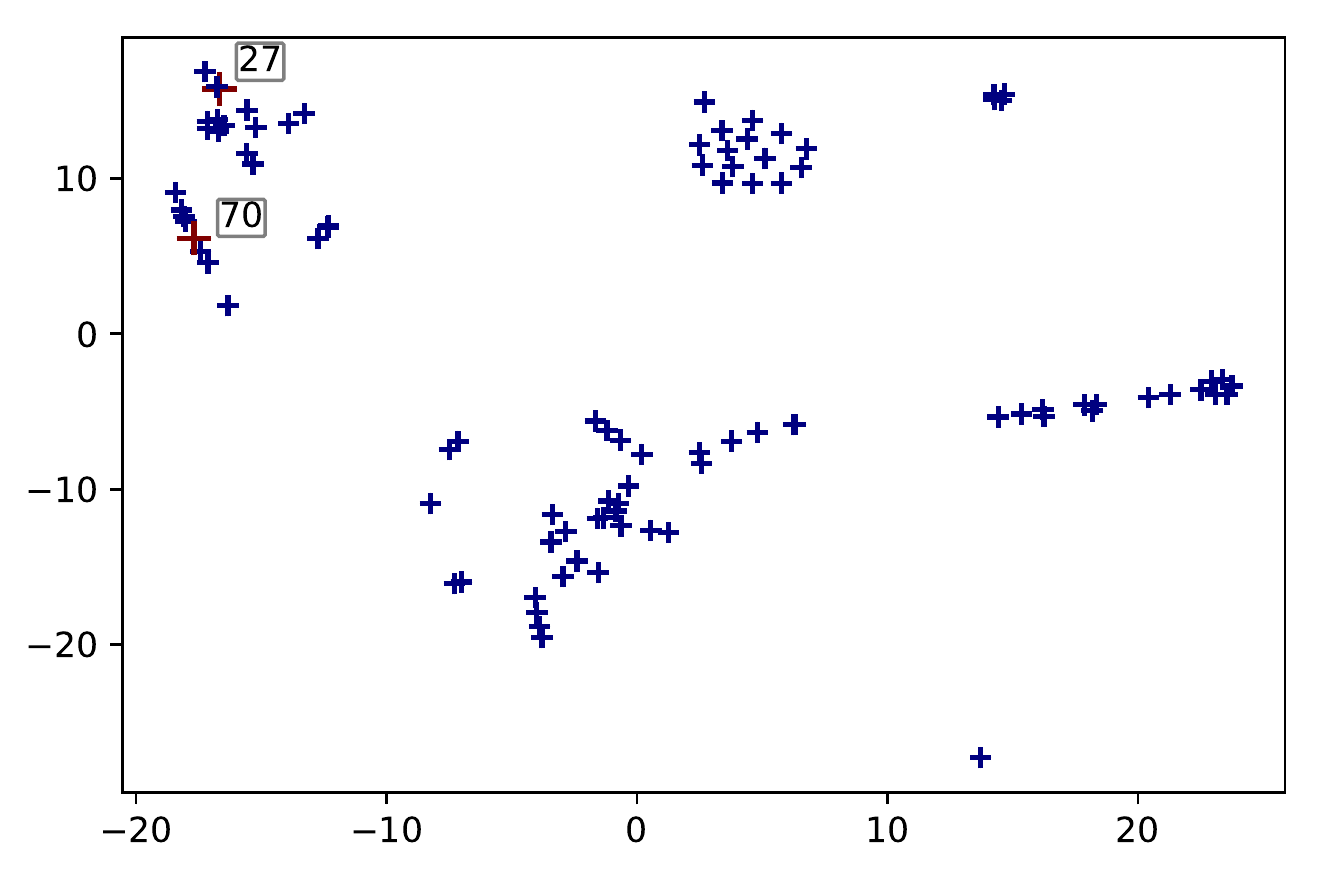}\\
(a) Train End Time & (b) Test Slot 1 End Time. 
\end{tabular}
\vspace{-2mm}
\caption{Use Case II. {\bf Top row:} GraphSage Embeddings. {\bf Bottom Row:} DyRep Embeddings. }
\label{fig:uc2}
\vspace{-2mm}
\end{figure*}

\item {\bf Nodes that have association but less number of events.} Nodes 19 and 26 remain associated with each other throughout data. DyRep Embeddings keep the nodes nearby although not in same cluster with a cosine distance of 0.649 which demonstrates its ability to learn the association and less communication dynamics between two nodes. For GraphSage the embeddings are on opposite ends of cluster with cosine distance of 1.964 Figure~\ref{fig:uc3} shows the visualization of embeddings at the two time points in both the methods.

\begin{figure*}[h!]
\small
\centering
\begin{tabular}{cc}
\includegraphics[width = 0.45\textwidth]{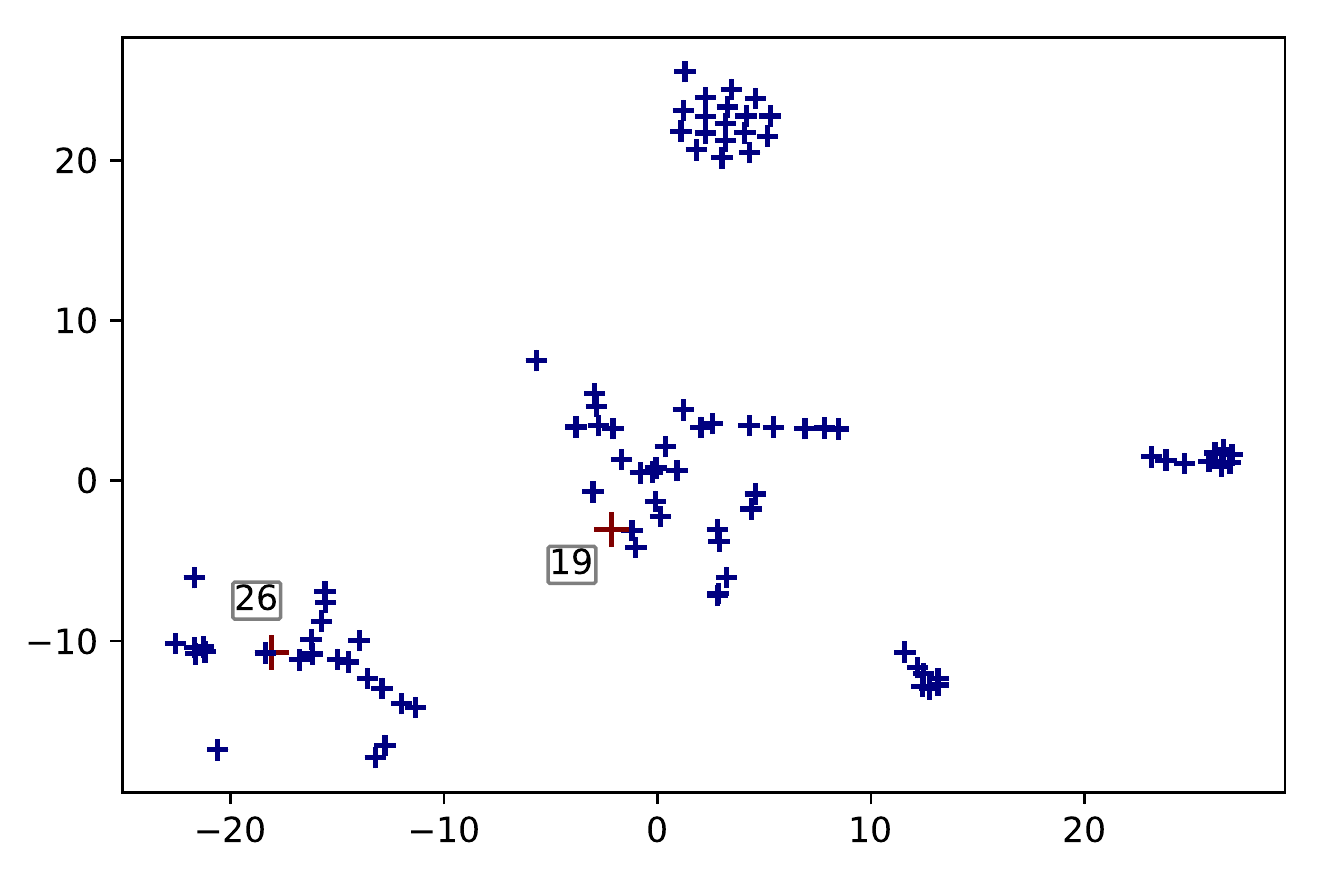}
& \includegraphics[width = 0.45\textwidth]{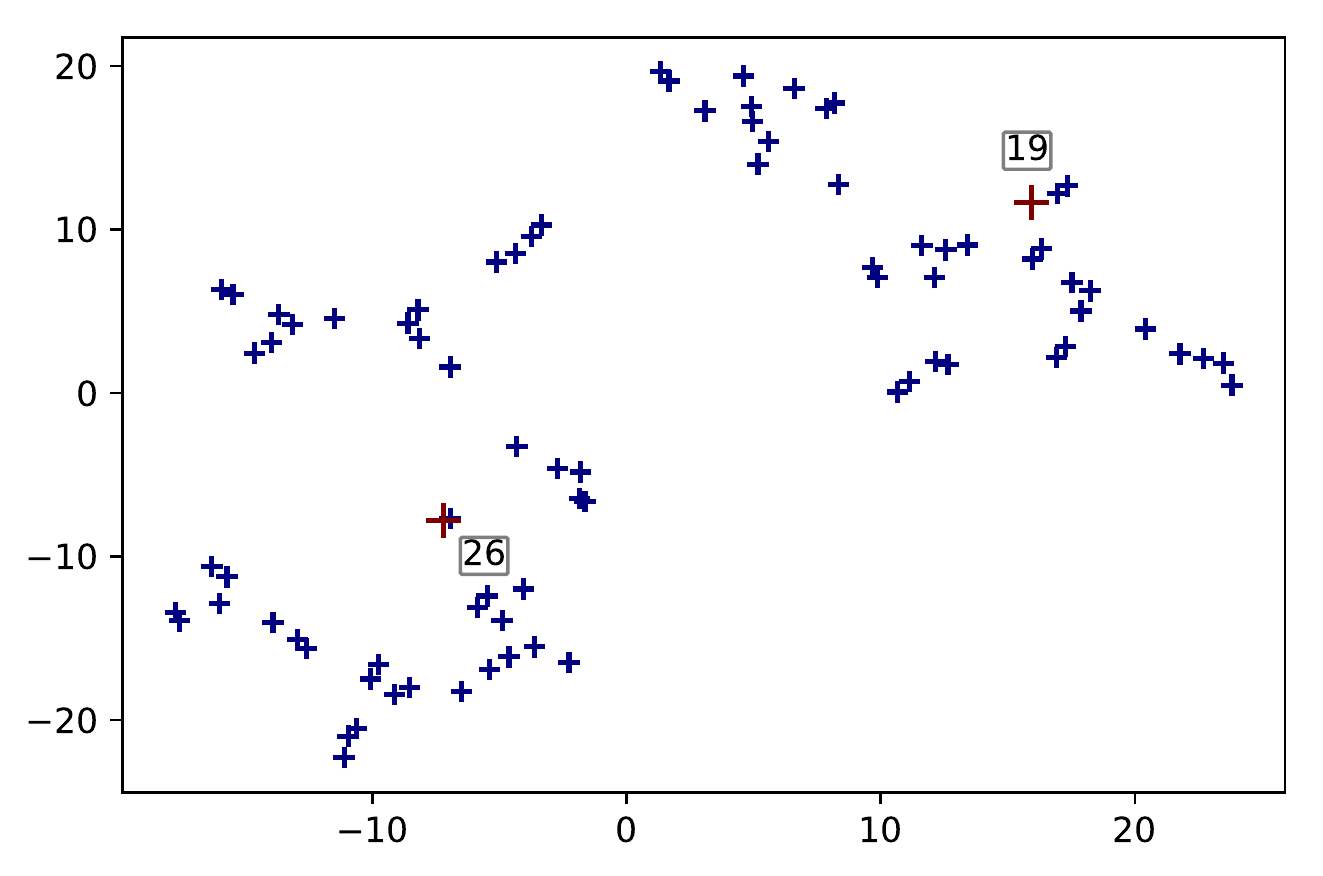}\\
(a) DyRep Embeddings & (b) GraphSage Embeddings. 
\end{tabular}
\vspace{-2mm}
\caption{Embedding Use Case III}
\label{fig:uc3}
\vspace{-2mm}
\end{figure*}

\item {\bf Temporal evolution of DyRep embeddings.} Here we visualize the embedding positions of the nodes (tracked in red) as they evolve through time and forms and breaks from  clusters.

\begin{figure*}[h!]
\small
\centering
\begin{tabular}{cc}
\includegraphics[width = 0.45\textwidth]{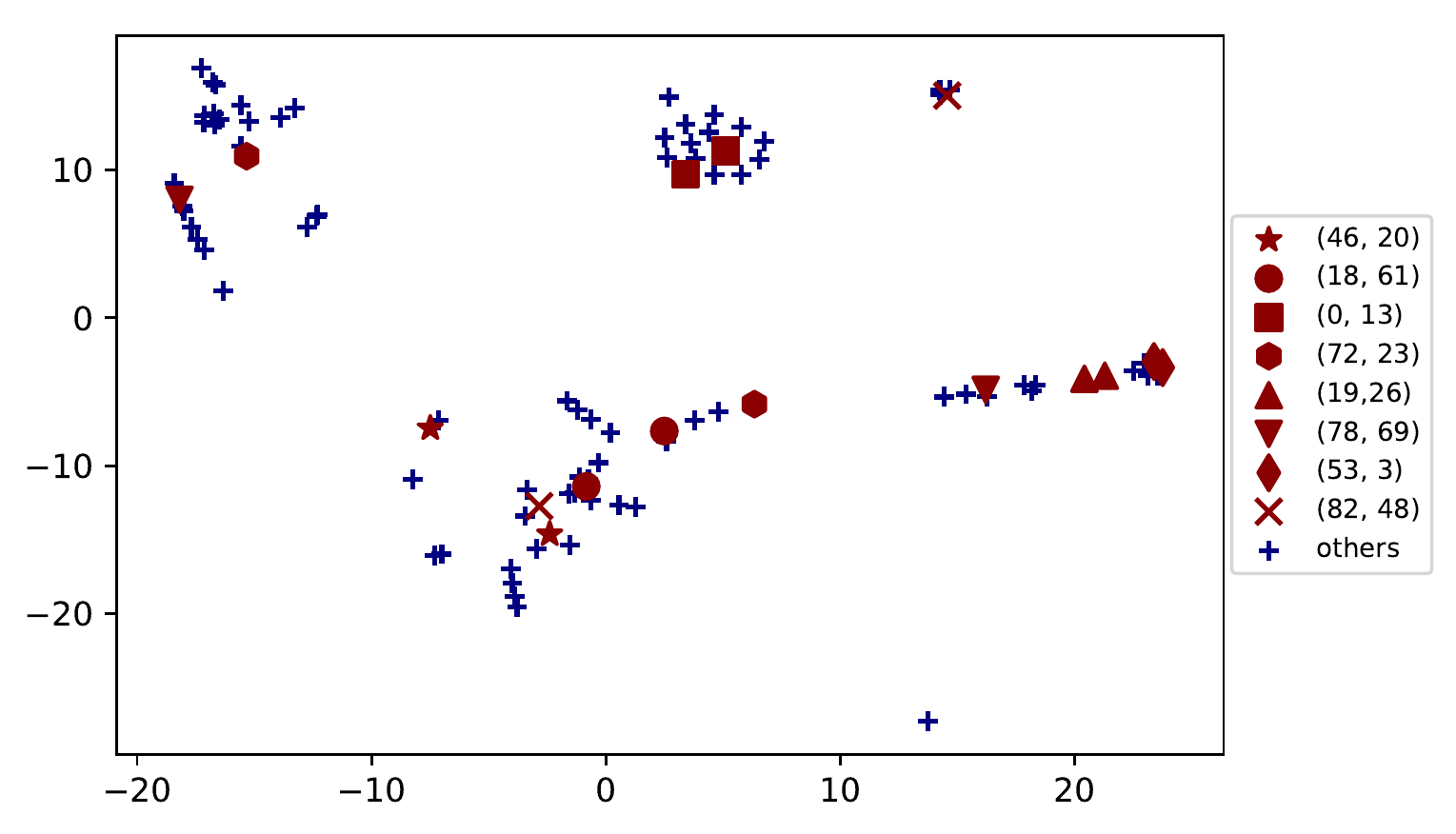}
& \includegraphics[width = 0.45\textwidth]{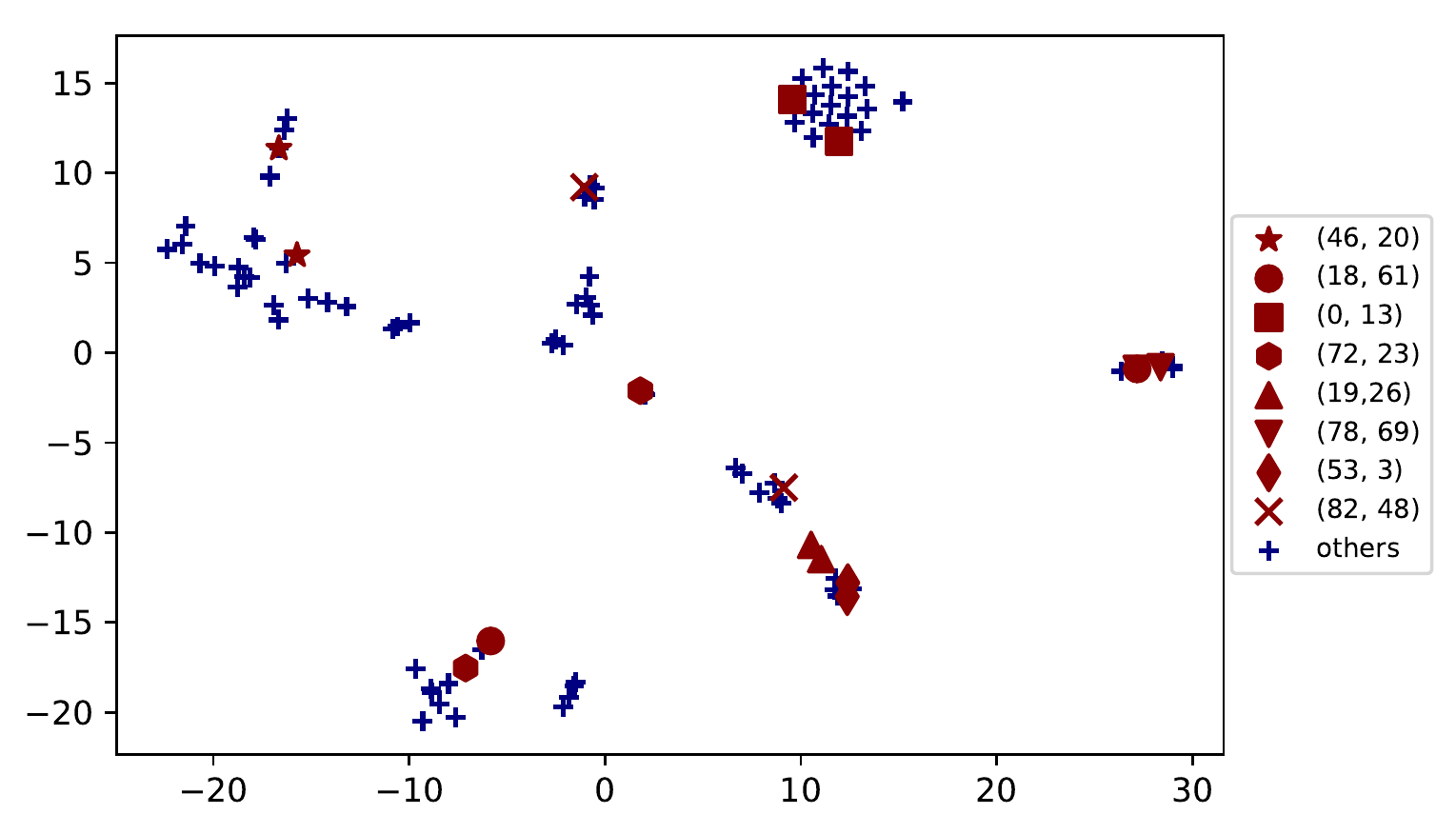}\\
$t=1$&$t=2$\\
\includegraphics[width = 0.45\textwidth]{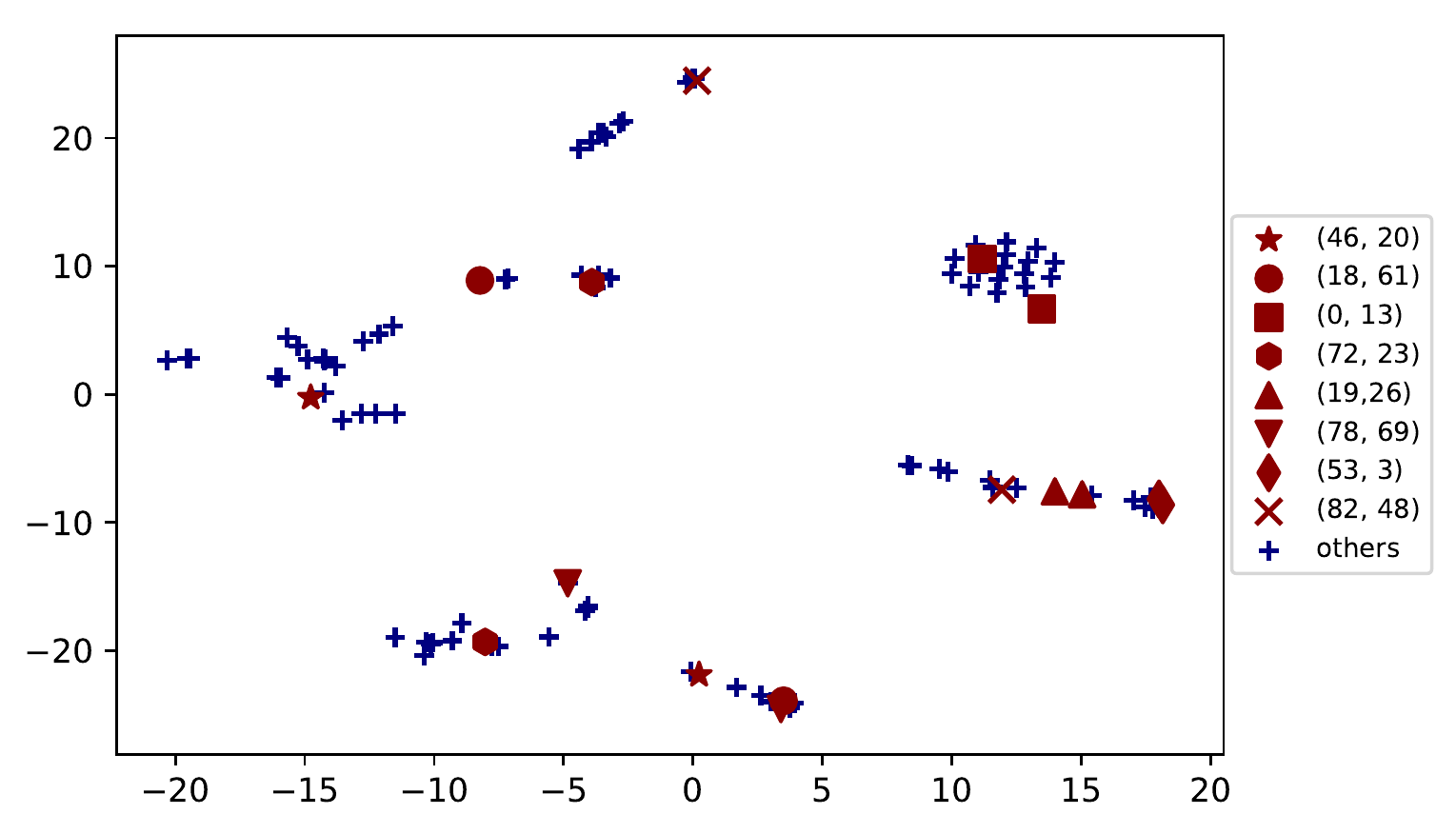}
& \includegraphics[width = 0.45\textwidth]{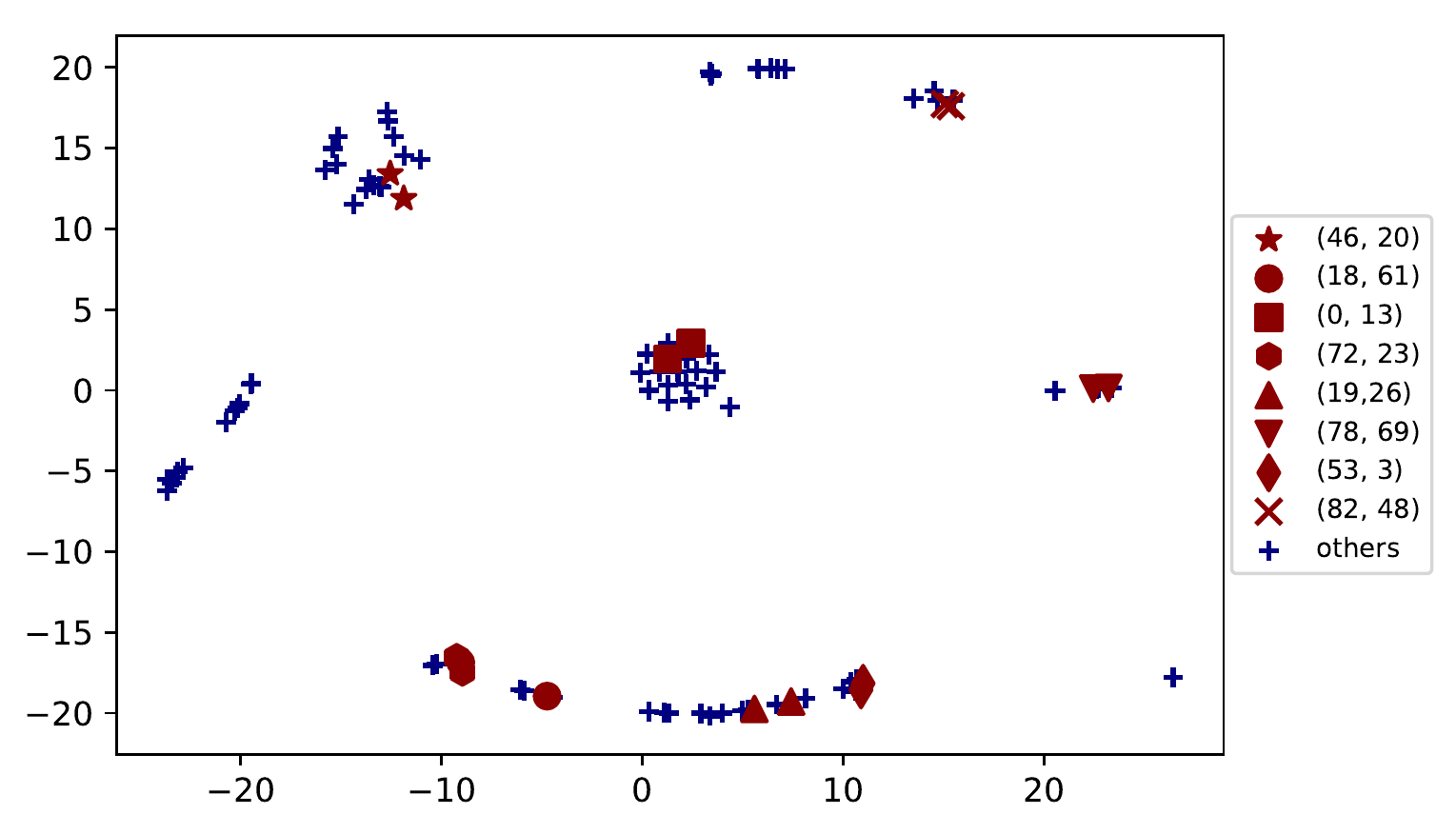}\\
$t=3$&$t=4$\\
\includegraphics[width = 0.45\textwidth]{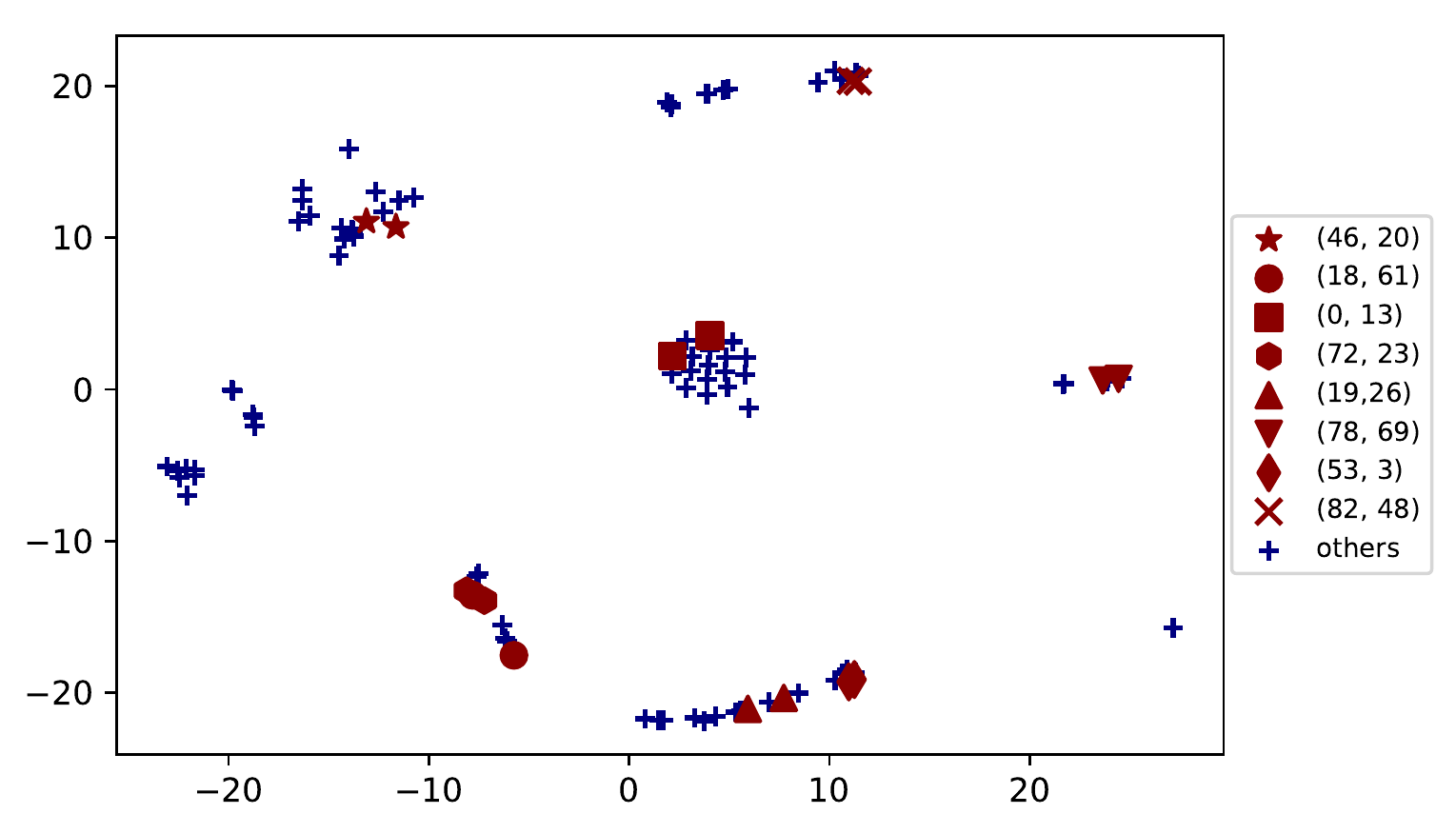}
& \includegraphics[width = 0.45\textwidth]{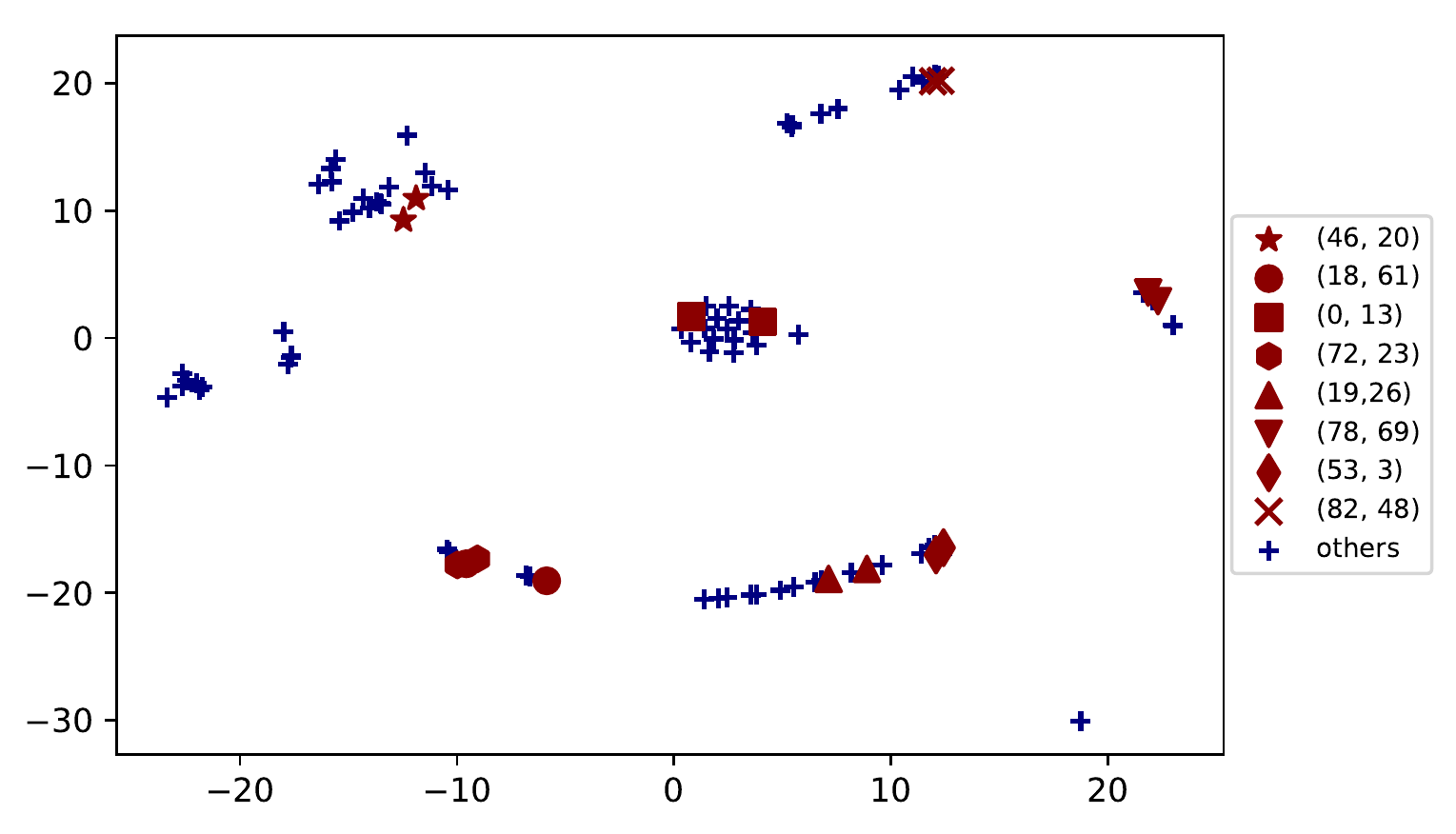}\\
$t=5$&$t=6$ 
\end{tabular}
\vspace{-2mm}
\caption{Use Case IV: DyRep Embeddings over time - From left to right and top to bottom. $t$ are the timepoints when test with that id ended. Hence, $t=1$ means the time when test slot 1 finished.}
\label{fig:uc4}
\vspace{-2mm}
\end{figure*}
\end{itemize}

\vspace{-2mm}
\section{Related Work}
\vspace{-2mm}

\textbf{Static Embedding Approaches.}
Representation learning over graph structured data has gained significant attention in recent years, specifically with the advent of deep learning as it provides many sophisticated techniques to capture various properties of graphs. Due to availability of data and lot of open research directions, most works in this area has focused on static graphs. Such approaches can be broadly classified into two categories -- Node embedding approaches aim to encode structural information pertaining to a node to produce its low-dimensional representation~\cite{CaoLuXu15,GroLes16,PerAlSki14,TanQuWanZhaetal15,WanCuiZhu16,WanCuiWanPeiZhuYan17,XuWeiCaoYu17}. As they learn each individual node's representation, they are inherently transductive. Recently,~\cite{HamYinLes17}  proposed GraphSage, an inductive method for learning functions to compute node representations that can be generalized to unseen nodes. Sub-graph embedding techniques learn to encode higher order graph structures into low dimensional vector representations~\cite{ScaGorTsoHagetal09, LiTarBroZem15, DaiDaiSon16}. Further, various approaches to use convolutional neural networks~\cite{KipWel16, KipWel16b, BruZarSzlLeC13} over graphs have been proposed to capture sophisticated feature information but are generally less scalable. Most of these approaches only work with static graphs or can model evolving graphs without temporal part.\\\\
\textbf{Dynamic Embedding Approaches.}
Preliminary approaches in dynamic representation learning have considered the two processes of communication and association in a segregated manner.~\cite{GoyKamHeLiu17} uses a warm start method to train across snapshots and employs a heuristic approach to learn stable embeddings over time but do not model time. Recently, Know-Evolve~\cite{TriDaiWanSon17} proposed a deep recurrent architecture to model multi-relational timestamped edges that addresses the communication process. This approach can be seen as addressing one part of the DyRep loop that we presented in Figure~\ref{fig:dyrep} and the information captured by node embeddings only depend on the edge-level information. DANE~\cite{LiDanHuTanChaLiu17} proposes a network embedding method in dynamic environment but their dynamics consists of change in node's attributes over time and their current work can be considered orthogonal to our approach. Research on learning dynamic embeddings has also progressed in language community where the aim is to learn temporally evolving word embeddings~\cite{BamMan17, RudBle18}. ~\cite{YanLiuWanLiuHan17, SarSidGor07} include some other approaches that propose model of learning dynamic embeddings in graph data but none of these models consider time at finer level and do not capture both topological evolution and interactions.\\\\
{\bf Deep Temporal Point Process Models.} Recently,~\cite{DuDaiTriUpaGomSon16} has shown that fixed parametric form of point processes lead into the model misspecification issues ultimately affecting performance on real world datasets. \cite{DuDaiTriUpaGomSon16} therefore propose a data driven alternative to instead learn the conditional intensity function from the observed events and thereby increase its flexibility. Following that work, there have been increased attraction in topic of learning conditional intensity function using deep learning\cite{MeiEis17} and also intensity free approach using GANS \cite{XiaFarYeYanYanSonZha17} for learning with deep generative temporal point process models.
\vspace{-2mm}
\section{Conclusion}
\vspace{-2mm}
We have presented a novel deep representation learning 
framework that can effectively and efficiently learn to compute 
node representations for dynamic graphs. We identified key challenges in such 
settings and devise an inductive approach to address them. 
We propose that node representations serve as mediator
that governs the complex and nonlinearly evolving processes of communication
and association over dynamic graphs. Our framework learns a set of functions that
can capture evolving dynamics of these processes and produce temporal 
and structural information-rich embeddings. Our superior predictive and qualitative evaluation 
performance demonstrates the effectiveness of our approach and we hope that
this contribution will open a wide range of application domains and exciting research directions in the area of representation learning
over dynamic graph structured data. Future interesting directions would be to extend DyRep to support network shrinkage settings (i.e. support node and edge deletions) and support encoding higher order dynamic structures.   
\newpage
\bibliography{bibfile}
\bibliographystyle{unsrt}

\end{document}